\documentclass[10pt,journal,compsoc]{IEEEtran} 
\usepackage{times}
\usepackage{epsfig}
\usepackage{graphicx}
\usepackage{amsthm}
\usepackage{amssymb}
\usepackage{multirow}

\usepackage{color}
\usepackage{subfigure}
\newcommand{\rc}[1]{\textcolor{red}{#1}}
\newcommand{\bc}[1]{\textcolor{blue}{#1}}

\newcommand{\mathvc}[1]{\mbox{\boldmath$#1$}}
\newcommand{\smathvc}[1]{\mbox{\footnotesize \boldmath$#1$}}

\newcommand{\argmin}{\mathop{\rm arg~min}\limits}
\newcommand{\eg}{{\itshape e.g.,~}}
\newcommand{\ie}{{\itshape i.e.,~}}
\newcommand{\etal}{{\itshape et al.~}}

\ifCLASSOPTIONcompsoc
\usepackage[nocompress]{cite}
\else
\usepackage{cite}
\fi
\ifCLASSINFOpdf
\else
\fi
\begin{document}
	%
	\title{Hierarchical Gaussian Descriptors with Application to Person Re-Identification}
	%
	%
	%
	%
	
	\author{Tetsu~Matsukawa,~\IEEEmembership{Member,~IEEE,}
		Takahiro~Okabe,~\IEEEmembership{Member,~IEEE,} \\
		Einoshin~Suzuki,~\IEEEmembership{Non~Member,~IEEE}
		and~Yoichi~Sato,~\IEEEmembership{Member,~IEEE}
		\IEEEcompsocitemizethanks{\IEEEcompsocthanksitem 
			T. Matsukawa and E. Suzuki are with the Faculty of Information Science and Electrical Engineering, Kyushu University, Fukuoka, Japan.
			E-mail: \{matsukawa, suzuki\}@inf.kyushu-u.ac.jp
			\IEEEcompsocthanksitem T. Okabe is with the Graduate School of Computer Science and Systems
			Engineering, Kyushu Institute of Technology, Fukuoka, Japan. 
			E-mail: okabe@ai.kyutech.ac.jp 
			\IEEEcompsocthanksitem Y. Sato is with the Institute of Industrial Science, The University of Tokyo, Tokyo, Japan. 
			E-mail: ysato@iis.u-tokyo.ac.jp  \protect\\
			Corresponding author: T.Matsukawa. }
		\thanks{This work was supported by the ``R\&D Program for Implementation of Anti-Crime and Anti-Terrorism Technologies for a Safe and Secure Society,'' under the fund for the integrated promotion of social system reform and research and development of MEXT Japan, and JSPS KAKENHI 15K16028.
		}}
\IEEEtitleabstractindextext{%
\begin{abstract}
Describing the color and textural information of a person image is one of the most crucial aspects of person re-identification (re-id). In this paper, we present novel meta-descriptors based on a hierarchical distribution of pixel features. Although hierarchical covariance descriptors have been successfully applied to image classification, the mean information of pixel features, which is absent from the covariance, tends to be the major discriminative information for person re-id. 
To solve this problem, we describe a local region in an image via hierarchical Gaussian distribution in which both means and covariances are included in their parameters. 
More specifically, the region is modeled as a set of multiple Gaussian distributions in which each Gaussian represents the appearance of a local patch. The characteristics of the set of Gaussians are again described by another Gaussian distribution. In both steps, we embed the parameters of the Gaussian into a point of Symmetric Positive Definite (SPD) matrix manifold. 
By changing the way to handle mean information in this embedding, we develop two hierarchical Gaussian descriptors. 
Additionally, we develop feature norm normalization methods with the ability to alleviate the biased trends that exist on the descriptors.
The experimental results conducted on five public datasets indicate that the proposed descriptors achieve remarkably high performance on person re-id.
\end{abstract}
			
}
			
			\maketitle

			\IEEEdisplaynontitleabstractindextext

			%
			\IEEEpeerreviewmaketitle
			
			\IEEEraisesectionheading{\section{Introduction}\label{sec:introduction}}
			%
			%
			%
			%
			
\IEEEPARstart{A}{ppearance} matching of person images observed in disjoint camera views, referred to as person re-identification (re-id), is receiving increasing attention, mainly because of its broad range of applications~\cite{Gong14, Zheng16, Karanam16}. 
In this task, the person images are captured from various viewpoints and under different illuminations, resolutions, human poses, and against various background environments.
These large intra-personal variations in person images cause considerable difficulties during attempts to match the person. In addition, similar clothes among different persons add further challenges.

Person images are low in resolution and are further characterized by large pose variations; consequently, it has been proven that the most important cue for person re-id is color information such as color histograms and color name descriptors~\cite{YangYYLYL14}. Because they cannot sufficiently differentiate different persons of similar colors, textural descriptors such as a Local Binary Pattern (LBP) and the responses of filter banks are often combined with color descriptors~\cite{RothHKBB14, xiong, zheng13}.
			\begin{figure}[t]
				\begin{center}
					 \includegraphics[width=0.99 \linewidth]{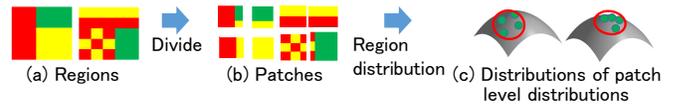}
				\end{center}
				\vspace{-6mm}
				\caption{Importance of hierarchal distribution: (a) Regions that have the same distribution (mean/covariance) of pixel features (each color indicates the same feature vector). (b) Local patches with a different pixel feature distribution inside the regions. (c) Regions can be distinguished via distributions of patch level distributions.}
				\label{fig:intution}
				\vspace{-4mm}
			\end{figure}

A covariance descriptor~\cite{TuzelPM06} describes a region of interest as a covariance matrix of pixel features. It provides a natural way to fuse different modalities, \eg the color and texture, of pixel features into a single meta-descriptor. Since the covariance matrix is obtained by averaging the features inside the region, it remedies the effects of noise and spatial misalignments. 
The covariance matrix lies on a specific type of Riemannian manifold, the manifold of Symmetric Positive Definite (SPD) matrices. In recent years, several metrics on the SPD matrix manifold were developed~\cite{Pennec2006, ArsignyFPA06}, and they provide grounding for the matching.
Consequently, the covariance descriptor has been successfully applied to person re-id~\cite{BakCCBT12, BakCBT12}.

In this paper, we propose novel meta-descriptors based on a hierarchical Gaussian distribution of pixel features. More specifically, our descriptors densely extract local patches inside a region and regard the region as a set of local patches. The region is firstly modeled as a set of multiple Gaussian distributions, each of which represents the appearance of one local patch. We refer to such a Gaussian distribution representing each local patch as a {\itshape patch Gaussian}. The characteristics of the set of patch Gaussians are again described by another Gaussian distribution. We refer to this Gaussian distribution as a {\itshape region Gaussian}. In both steps, we embed parameters of one Gaussian distribution into a point on the SPD matrix manifold.

Our motivation for the use of a hierarchical distribution stems from the structural appearance of person images. A person's clothes consist of local parts, each of which has local color/texture structures. The spatial arrangement of these parts determines the global structural appearance. However, most of the existing meta-descriptors~\cite{CarreiraCBS15, GongWL09, Ma14, Nakayama10, Serra15, TuzelPM06} are based on a global distribution of pixel features inside a region and thus the local structure of the person's image is lost. In contrast, we describe the global distribution using the local distribution of the pixel features. This enables us to distinguish textures with the same global distribution but different local structures, as illustrated in Fig.~ \ref{fig:intution}.

\begin{figure}[t]
				\begin{center}
					\includegraphics[width=1.0 \linewidth]{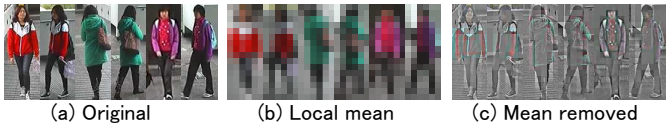}
					\vspace{-11mm}
				\end{center}
				\caption{Importance of mean: (a) Original images. (b) Images that show mean RGB values of $10\times10$ pixel patches of (a). (c) Mean removed images (each RGB value is scaled over the range [0,255] for visualization). It is easy to determine the same persons from (b), whereas it is difficult from (c).}
				\label{fig:RGB}
				\vspace{-4mm}
			\end{figure}

We use the Gaussian distribution as a base component of the hierarchy. The motivation of the use of this distribution originates from the importance of the mean color of local parts.
Although hierarchical covariance representations have been proposed~\cite{LiW12, serra14}, the mean information is not included in each hierarchy. 
The absence of the mean information is crucial when applied to person re-id. This is because the clothes a person wears tend to consist of a small number of colors in each local part, and therefore the mean color in the local parts tends to be the major information that makes it possible to discriminate among persons.
As shown in Fig.~\ref{fig:RGB}, the mean images of local color contain highly distinguishing information of different persons.
		
We name the proposed meta-descriptors {\bf H}{\itshape ierarchical} {\bf G}{\itshape aussian} {\bf D}{\itshape escriptors} ({\bf HGD}s). 
The main contributions of this paper\footnotemark~are summarized below:

\footnotetext{This paper is a substantially extended version of our paper in the CVPR2016 proceedings~\cite{Matsukawa16}. The HGDs refer to the generic name including the {\bf G}{\itshape aussian} {\bf O}{\itshape f} {\bf G}{\itshape aussians} ({\bf GOG}) descriptor proposed in~\cite{Matsukawa16} and its simplification, {\bf Z}{\itshape ero mean Gaussian} {\bf O}{\itshape f} {\bf Z}{\itshape ero mean Gaussians} ({\bf ZOZ}) descriptor.
}

\begin{enumerate}
\setlength{\leftskip}{-0.5cm}
\item We present effective hand-crafted descriptors for person re-id. The HGDs provide a conceptually simple and consistent way to generate discriminative and robust features that describe color and textural information simultaneously.
\item We propose the hierarchical use of Gaussian embedding of pixel features ({\bf GOG}). We experimentally validate the importance of both the hierarchical distribution and mean information of pixel features for person re-id.
\item We define a scale normalization of an SPD matrix and validate its importance for HGDs. Based on the scale normalization, we develop a new zero mean Gaussian embedding. The HGDs based on this embedding can achieve close performance to the HGDs using the original Gaussian embedding of Lovri{\'c} \etal~\cite{lovric2000} with smaller dimensionality and computational cost ({\bf ZOZ}).
\item For normalizing the norm of HGDs, we point out a biased trend of the SPD matrices on Log-Euclidean (LE) tangent space. To alleviate this effect, we propose norm normalization methods accompanied with mean removal on the SPD matrix manifold. In~\cite{Matsukawa16}, we validated the effectiveness of the mean removal before L2 norm normalization. In this paper, we extend this normalization with the intrinsic properties of the Riemannian manifold.
\end{enumerate}

\section{Related Work}
{\bf Meta-descriptors of local features.}
By {\itshape embedding} the probability distribution into a matrix, a meta-descriptor such as the covariance descriptor summarizes the local features inside a region.
Several methods, such as Entropy and Mutual Information matrix~\cite{Biagio}, Brownian covariance matrix~\cite{Bak16}, and covariance matrix on Reproducing Kernel Hilbert Space (RKHS)~\cite{Harandi14, Minh14} were proposed to improve the description ability of the covariance descriptor. Gaussian descriptors, such as Shape of Gaussians~\cite{GongWL09}, Global Gaussian~\cite{Nakayama10} and Gaussians Of Local Descriptors (GOLD)~\cite{Serra15}, embed both the mean vector and covariance matrix of local features. Second-order Average and Max Poolings (2AvgP and 2MaxP)~\cite{CarreiraCBS15, Hong16} summarize the pairwise correlation of local feature components without subtracting the mean information. The improved Gaussian embedding~\cite{Wang16b, Wang16} adjusts the contributions of the mean vector and covariance matrix in the embedding.
These representations are sometimes used for training-free alternative of {\itshape coding}-based summaries, such as Bag-Of-Words (BOW)~\cite{Csurka04}, Fisher Vector (FV)~\cite{SanchezPMV13}, and Vectors of Locally Aggregated Descriptors (VLAD)~\cite{JegouPDSPS12}.
			
Recently, several hierarchal meta-descriptors, which construct descriptors over local covariance matrices, have been proposed.
Li \etal proposed a Local Log-Euclidean Covariance Matrix (L$^2$ECM), which is a vector map obtained by mapping the local covariance matrices into LE tangent space~\cite{LiW12}.
They demonstrated applications of the statistical modeling of L$^2$ECM by the second-order moment (covariance matrix) in human detection, texture classification, and object tracking. 
Faraki \etal proposed BOW, FV, and VLAD codings on covariance matrices~\cite{Faraki15PR, Faraki15}. 
S{\'{a}}nchez \etal extended Gaussian FV coding to a broader family of distribution, including Wishart distribution, which is a distribution of covariance matrices~\cite{Sanchez15}.
Ilea \etal proposed a Riemannian FV coding based on the mixture model of Riemannian Gaussian distributions~\cite{Ilea16}.
Most importantly, previous descriptors adopt local covariance matrices for the input of another coding/embedding.

Concurrent with our initial version~\cite{Matsukawa16}, Li \etal extended L$^2$ECM to Local Log-Euclidean Multivariate Gaussian (L$^2$EMG)~\cite{Li16}, which is a vector map of local Gaussians. They used L$^2$EMG as the input of FV coding and applied it to image classification. 
Codebook free descriptors are more adequate for person re-id because this task needs to match people that are not necessarily included in the training dataset from which the codebook is learned.
			
\noindent{\bf Scale normalization of SPD matrix for LERM}.
The Log-Euclidean Riemannian Metric (LERM)~\cite{ArsignyFPA06} is widely used as a metric for SPD matrix manifolds. The most popular application of LERM is for calculating the distance or mean of SPD matrices. Several authors developed modifications of LERM for improved processing of the manifold, \eg the Campbell-Baker-Hausdorff (CBH) expansion~\cite{Tosato13} and LERM for infinite-dimensional covariance matrices ~\cite{Minh14}. However, we do not see any discussion on the scale effect of the SPD matrix on LE tangent space. This prompted us to address the biased diagonal trend of matrices on LE tangent space and overcome this problem by the scale normalization.
			
\noindent{\bf Norm normalization of SPD matrix descriptor.} 
To date, norm normalizations such as L1, L1-sqrt, and L2 norm normalizations were intensively applied to histogram-based feature representations, such as SIFT and BOW~\cite{Kobayashi14, Xie16}. In particular, S{\'{a}}nchez \etal argued that norm normalization can help to improve the recognition accuracies of any high-dimensional features~\cite{SanchezPMV13}. Regarding the feature vectors of the SPD matrix descriptors, we rarely see discussions of norm normalization. 
In this paper, we point out that feature vectors on LE tangent space can be largely biased because the origin of this space is an identity matrix. 
We solve this problem by using the tangent space of the Riemannian mean of the training data rather than the identify matrix. Although this space is sometimes used for improving the distance calculation~\cite{TuzelPM08, barachant2013}, we validate its effectiveness for norm normalization.  
			
\noindent{\bf Feature representation in person re-id. }
Several feature representations have been developed for unsupervised person re-id by focusing on the unique properties of person images.
Symmetry-Driven Accumulation of Local Features (SDALF)~\cite{bazzani} exploits the symmetric property of a person by obtaining their head, torso, and leg positions to overcome view variations. A color-invariant signature robustly describes color information under the different illumination conditions~\cite{kviatkovsky2013}. 
Unsupervised salience learning~\cite{ZhaoOW13, ZhaoOW16} estimates rare patches among different images to match rare appearances such as rare-colored coats, baggage, and folders. Attribute based descriptors obtain a lingual description of person images~\cite{layne2012person}.

In recent years, a supervised approach, \ie metric learning, has shown more impressive results in terms of accuracies~\cite{paisi, PedagadiOVB13, RothHKBB14, zheng13, xiong, Liao15, Zhang16}. 
The features used for metric learning are rather simple compared to the features for unsupervised settings. Metric learning requires features to contain sufficient information. For example, high-dimensional features composed of densely sampled color histograms are often processed by using LBPs and SIFTs~\cite{PedagadiOVB13, xiong}. 
The design of features would largely affect the matching accuracy of metric learning methods. 
Nevertheless, most of the previous work focused on an algorithm of metric learning~\cite{RothHKBB14, zheng13}, and comparatively little research has been devoted to studying the feature design~\cite{Liao15, MaSJ12, YangYYLYL14}. Extensive reviews of recent progress on person re-id are found in~\cite{Zheng16, Karanam16}.

Our use of two-level (patch/region) statistics for the re-id descriptor was to some extent motivated by Local Maximal Occurrence (LOMO)~\cite{Liao15}. This method locally constructs a histogram of pixel features, and takes its maximum values within horizontal strips to overcome viewpoint variations while maintaining local discrimination. Indeed, LOMO only describes the mean information of pixel features. 
Although the use of the mean and covariance information via Gaussian embedding was introduced into person re-id~\cite{Ma14}, it was not constructed in a hierarchal manner.

A Convolutional Neural Network (CNN) is one of the state-of-the art recognition algorithms that leverages a hierarchal structure~\cite{krizhevsky}. 
CNN has been gradually improving the accuracy of person re-id~\cite{LiZXW14, ahmed15, Cheng16, Varior16a, Subramaniam16}. However, CNN relies on a large number of labeled training samples. 
In small-sampled datasets such as the VIPeR dataset~\cite{Gray}, the performance of the traditional metric learning approach continues to be competitive with CNN. 
In contrast to CNN, our descriptors require no training samples (except for optional norm normalization) because in each hierarchy, our descriptor describes the regions via mean and covariance estimations on features of an input image. 
\section{Background Theory}
In this section, we explain the background theory by focusing on the SPD matrix manifold. 
First, we briefly review the Riemannian geometry of the SPD matrices. More in-depth treatment of manifolds and related topics can be found in~\cite{Liu12, Turage14}. 
We then define a scale normalization, which will be adopted for the new Gaussian embedding described in the following section.

\subsection{Riemannian Geometry of SPD Matrices}
\label{sec:LE}
A Riemannian manifold $\mathcal{M}$ is a smooth manifold equipped with a Riemannian metric, \ie a smoothly varying inner product on its tangent space $T_p \mathcal{M}, p \in \mathcal{M}$. 
The minimum length curve connecting two points on the manifold is known as the geodesic, and the distance between the points is given by the length of the curve. 
Two operators, known as the exponential map {\rm exp}$_{p}: T_p\mathcal{M} \rightarrow \mathcal{M}$ and the logarithmic map {\rm log}$_{p}: \mathcal{M} \rightarrow T_p \mathcal{M}$ switch between the manifold and the tangent space at $p$.
We consider the special type of manifold $\mathcal{M}$, the manifold of $d \times d$ SPD matrix $Sym_{d}^{+}$. 
			
In the Affine Invariant Riemannian Metric (AIRM)~\cite{Pennec2006}, the exponential map ${\rm exp}_{\smathvc{P}}: T_{\smathvc{P}} Sym_{d}^{+} \rightarrow Sym_{d}^{+}$ and 
the logrithmic map ${\rm log}_{\smathvc{P}}: Sym_{d}^{+} \rightarrow T_{\smathvc{P}} Sym_{d}^{+}$ at the point \mathvc{P} $\in Sym_{d}^{+}$ are respectively defined as: 
\begin{eqnarray}
{\rm exp}_{\smathvc{P}} \mathvc{X} &=& \mathvc{P}^{\frac{1}{2}} {\rm exp}\left( \mathvc{P}^{-\frac{1}{2}} \mathvc{X} \mathvc{P}^{-\frac{1}{2}} \right) \mathvc{P}^{\frac{1}{2}}, \\
{\rm log}_{\smathvc{P}} \mathvc{X} &=& \mathvc{P}^{\frac{1}{2}} {\rm log}\left( \mathvc{P}^{-\frac{1}{2}} \mathvc{X} \mathvc{P}^{-\frac{1}{2}} \right) \mathvc{P}^{\frac{1}{2}}. \label{eq:logmapAIRM} 
\end{eqnarray}
Here {\rm log}$\left( \cdot \right)$ and {\rm exp}$\left( \cdot \right)$ are the principal matrix logarithm and exponential operator, respectively. 
Let $\mathvc{X} = \mathvc{U} {\rm Diag}(\lambda_i) \mathvc{U}^T$ be the eigen decomposition of a symmetric matrix \mathvc{X}. 
Here ${\rm Diag}(\lambda_i)$ is a diagonal matrix formed from the eigen values $\lambda_1,..., \lambda_d$ and \mathvc{U} is the eigen vectors \ie $\mathvc{U}^T\mathvc{U} = \mathvc{U}\mathvc{U}^T = \mathvc{I}_d$, where \mathvc{I}$_d$ is a $d$-dimensional identity matrix. 
The principal matrix exponential and logarithm of \mathvc{X}, respectively, are defined as,
\begin{eqnarray}
{\rm exp}\mathvc{X} &=& \mathvc{U} {\rm Diag} \left({\rm exp}\left(\lambda_i\right) \right) \mathvc{U}^T,\\
{\rm log}\mathvc{X} &=& \mathvc{U} {\rm Diag} \left({\rm ln}\left(\lambda_i\right) \right) \mathvc{U}^T \label{eq:logm}.
\end{eqnarray}
The logarithmic mapping in Eq.(\ref{eq:logmapAIRM}) relates to the geodesic on the manifold, and its norm coincides with the following geodesic distance between \mathvc{X}, \mathvc{Y} $\in Sym_d^{+}$:
\begin{equation}
D_{\mbox{\scriptsize AIRM}}(\mathvc{X}, \mathvc{Y}) = \Vert {\rm log}_{\smathvc{X}} \mathvc{Y} \Vert_F  = \sqrt{ {\rm Tr}\left({\rm log}^2 \left(\mathvc{X}^{-\frac{1}{2}} \mathvc{Y} \mathvc{X}^{-\frac{1}{2}} \right)\right)},
\label{eq:AIRM}
\end{equation}
where $\rm{Tr}(\cdot)$ denotes the trace norm of the matrix. 
        
The Log-Euclidean Riemannian Metric (LERM)~\cite{ArsignyFPA06} is another metric of SPD matrix manifolds. 
It is derived by leveraging the Lie group structure of $Sym^{+}_{d}$ under the group operation $\mathvc{X} \odot \mathvc{Y} := {\rm exp}\left({\rm log}\mathvc{X} + {\rm log}\mathvc{Y} \right)$.
In the LERM, the geodesic distance is defined as:
\begin{equation}
D_{\mbox{\scriptsize LERM}}(\mathvc{X}, \mathvc{Y} ) = \Vert {\rm log}\mathvc{X}  - {\rm log}\mathvc{Y} \Vert_F.
\end{equation}

The advantage of LERM is that it can decouple the distance calculation and tangent space mapping. 
In LERM, the tangent space mapping is performed through the principal matrix logarithm in Eq.(\ref{eq:logm}), which is a special case of the Eq.(\ref{eq:logmapAIRM}) when the tangent pole is the identity matrix, \ie $\mathvc{P}=\mathvc{I}_d$.

\begin{figure}[t]
				\begin{center}
                                        \includegraphics[width=0.9 \linewidth]{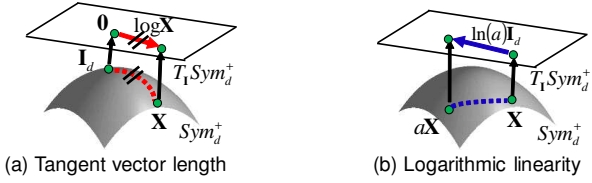}
					\vspace{-7mm}
				\end{center}
				\caption{ 
					Properties of LE tangent space: 
					(a) The length of the tangent vector is the geodesic distance from the identity matrix. 
					(b) Scaling the SPD matrix affects only diagonal elements in the tangent space.
					\label{fig:LEtangent}}
                               \vspace{-4mm}
			\end{figure}

For the convenience of the subsequent explanation, we point out the following two properties of LERM (Fig.~\ref{fig:LEtangent}).
			
\noindent {\bf Property 1 (Tangent vector length).} 
The mapped matrix ${\rm log}\mathvc{X}$ is the relation of the geodesic between the identity matrix and \mathvc{X}.
Its norm coincides with both geodesic distances $D_{\mbox{\scriptsize LERM}}$ and $D_{\mbox{\scriptsize AIRM}}$.
			\begin{proof} 
				Since ${\rm log} \mathvc{I}_d = \mathvc{0}$ and $\mathvc{I}_d^{\frac{1}{2}} = \mathvc{I}_d^{-\frac{1}{2}} = \mathvc{I}_d$, we can verify that 
				$D_{\mbox{\scriptsize LERM}}(\mathvc{I}_d, \mathvc{X} ) = D_{\mbox{\scriptsize AIRM}}(\mathvc{I}_d, \mathvc{X}) = \Vert {\rm log}\mathvc{X} \Vert_F$.
			\end{proof}
\noindent {\bf Property 2 (Logarithmic linearity).} 
LERM inherits the logarithmic linearity of the scalar space as follows. 
			\begin{equation}
			{\rm log}(a \mathvc{X}) = {\rm log}\mathvc{X} + {\rm ln}(a) \mathvc{I}_d. \label{eq:log}
			\end{equation}
                        \vspace{-1mm}
                        Here $a \in \mathbb{R}^1$ is an arbitrary scalar value that satisfies $a > 0$.
			\begin{proof} We can confirm that 
				$ {\rm log}(a \mathvc{X}) = \mathvc{U} {\rm Diag} ({\rm ln}(a \lambda_i) ) \mathvc{U}^T  
				= \mathvc{U} {\rm Diag} ({\rm ln}(\lambda_i) + {\rm ln}(a)) \mathvc{U}^T 
				= {\rm log}(\mathvc{X}) + {\rm ln}(a) \mathvc{U} \mathvc{U}^T  = {\rm log}(\mathvc{X}) + {\rm ln}(a) \mathvc{I}_d. $
			\end{proof}
                        
		\def\subfigcapskip{-5pt}  
\vspace{-4.5mm}
\subsection{Scale Normalization of SPD Matrix}
\label{sec:scalenorm}
We define a scale normalization of the SPD matrix with the ability to canonicalize the diagonal elements of SPD matrices on LE tangent space. 
We will see that this normalization is inherently adopted in the Gaussian embedding of Lovri{\'c} \etal~\cite{lovric2000} ($\S \ref{sec:patch}$). 
			
			\noindent {\bf Definition (Scale normalization).} Given an SPD matrix \mathvc{X}$\in Sym_d^{+}$, we define the scale normalization: $\eta: Sym_d^+ \rightarrow Sym_d^+$ as follows:
                        \begin{equation}
			\eta(\mathvc{X}) = |\mathvc{X}|^{-\frac{1}{d}} \mathvc{X}, \label{eq:scalenorm}
			\end{equation}
			where $|\cdot|$ is the determinant of a matrix.
			We derive the following properties for the scale normalization.
                        
			\noindent {\bf Property 1 (Scale invariance).} 
			For any positive scalar $a \in \mathbb{R}^1$, the scale normalization has the following invariance.
                        \vspace{-1mm}
			\begin{equation}
			\eta(a\mathvc{X}) = \eta(\mathvc{X}).
			\end{equation}
			\begin{proof}
				For any $d \times d$ matrix \mathvc{X} and scalar $a$, the following relation holds: $|a \mathvc{X}| = a^{d} |\mathvc{X}|$ ~\cite{Harville}.
				Therefore,
				$\eta(a\mathvc{X}) =  |a \mathvc{X}|^{-\frac{1}{d}} a \mathvc{X} \nonumber =  ( a^{d} |\mathvc{X}| )^{-\frac{1}{d}} a \mathvc{X} 
				=  a^{-1} |\mathvc{X}|^{-\frac{1}{d}} a \mathvc{X} =  |\mathvc{X}|^{-\frac{1}{d}} \mathvc{X} =  \eta(\mathvc{X}).$
			\end{proof}
			\vspace{-0.5mm}
			\noindent {\bf Property 2 (Canonical eigenvalues).} 
			The mean logarithmic eigenvalues of the $\eta(\mathvc{X})$ are canonicalized to zero.
			\begin{proof}
				Since $|\mathvc{X}|=\prod_{j=1}^{d}\lambda_j$, we have 
				$\rm{ln}(|\mathvc{X}|^{-\frac{1}{d}}) = \rm{ln}((\prod_{j=1}^{d}\lambda_j)^{-\frac{1}{d}}) = - \frac{1}{d} \rm{ln}(\prod_{j=1}^{d} \lambda_j) = - \frac{1}{d} \sum_{j=1}^d \rm{ln}\lambda_j$. 
				Therefore, using Eq.(\ref{eq:log}), we have $ \rm{log} \eta(\mathvc{X}) = \rm{log}\mathvc{X}  + \rm{ln}(|\mathvc{X}|^{-\frac{1}{d}}) \mathvc{I}_d 
				= \rm{log}\mathvc{X}  - \frac{1}{d} (\sum_{j=1}^d \rm{ln}\lambda_j) \mathvc{I}_d  = \mathvc{U} \rm{Diag}(\lambda_i - \frac{1}{d}\sum_{j=1}^d {\rm ln} \lambda_j)\mathvc{U}^T$.
			\end{proof}
			\noindent{\bf Direct tangent space mapping.}
			The scale normalization changes the eigenspectrum of $\mathvc{X}$ as $\lambda_i \leftarrow  \lambda_i/(\prod_{j=1}^{d}\lambda_j)^{\frac{1}{d}}, i=1,...,d$.
			When the dimensionality $d$ is high, many eigenvalues could be small \ie less than 1. In such a case, it causes division by zero because the denominator becomes too small due to repeated multiplication by small eigenvalues.
			We avoid this problem by directly calculating the matrix values on LE tangent space by: ${\rm log} \eta(\mathvc{X}) = {\rm log}\mathvc{X}  - \frac{1}{d} (\sum_{j=1}^d \rm{ln}\lambda_j) \mathvc{I}_d$.
			\vspace{-1mm}
			\section{Hierarchical Gaussian Descriptors }
			In this section, we propose two {\bf H}{\itshape ierarchical} {\bf G}{\itshape aussian} {\bf D}{\itshape escriptors} ({\bf HGD}s). 
			Both HGDs are based on a common pipeline that follows two motivations: 1) The hierarchical distribution is discriminative because it can distinguish textures with a similar global distribution but a different local distribution of pixel features (Fig.~\ref{fig:intution}). 2) When constructing the hierarchal descriptor, the mean information of pixel features in local parts tends to be major discriminative information (Fig.~\ref{fig:RGB}).
			
\begin{figure*}[t]
				\begin{center}
                                	\includegraphics[width=1 \linewidth]{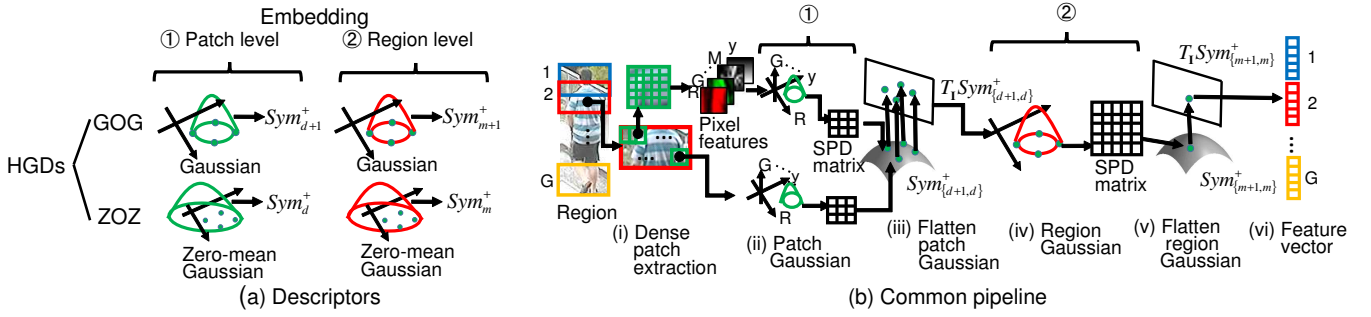}
                                        
					\vspace{-5mm}
				\end{center}
\caption{ 
{\bf Hierarchial Gaussian Descriptors (HGDs):} 
(a) Two descriptors that we extract by changing the Gaussian embeddings of patch/region levels. 
(b) Common pipeline for each descriptor:
(i) Densely extract local patches located inside each region.
(ii) Describe each of these local patches via a Gaussian distribution of pixel features which we refer to as a patch Gaussian.
(iii) Flatten and vectorize each of patch Gaussians by considering their underlying Riemannian geometry.
(iv) Summarize patch Gaussians inside a region into a region Gaussian. 
(v) Flatten the region Gaussian and create a feature vector.
(vi) Concatenate feature vectors extracted from all regions into one vector. 
\label{fig:flow}}
                        \vspace{-1mm}
			\end{figure*}

                        Two HGDs named {\bf G}{\itshape aussian} {\bf O}{\itshape f} {\bf G}{\itshape aussians} ({\bf GOG}) and {\bf Z}{\itshape ero mean Gaussian} {\bf O}{\itshape f} {\bf Z}{\itshape ero mean Gaussians} ({\bf ZOZ}) descriptors are extracted by changing the base Gaussian embedding of the common pipeline (Fig.~\ref{fig:flow}(a)). Each of the HGDs provides the features of regions in vector form to enable conventional metric learning methods to be easily applied.
                        			
                        We outline the common pipeline of HGDs in Fig.~\ref{fig:flow} (b). The feature representation of a person image is obtained by adopting a part-based model, which divides a person image into $G$ regions. The division of a person image is arbitrary, \eg the estimation of human body parts would be used. In this paper, we assume the $G$ regions are given in advance. In particular, we use fixed horizontal stripes to enhance the view invariance~\cite{Liao15}. For each region, we characterize each pixel by low-level features such as its color and gradient. We summarize them in a two-level (patch/region) hierarchical distribution. In each hierarchy, we summarize the feature distribution by one of the following two embeddings; the Lovri{\'c}'s Gaussian embedding~\cite{lovric2000} and the Zero mean Gaussian (ZmG) embedding, which we propose in $\S$ \ref{sec:zmg}. 
                        
			\def\subfigcapskip{-5pt} 
                        \vspace{-1mm}
			\subsection{Pixel Features}
			\label{sec:pixel}
			Let us focus on one of the $G$ regions of a person image. We describe the local structure of the region by densely extracting squared (k $\times$ k pixels) patches with $p$ pixel intervals (Fig.~\ref{fig:flow} (b-i)). In order to characterize each pixel in the patch, we extract a $d$-dimensional feature vector $\mathvc{f}_i$ for every pixel $i$. The feature vector can consist of any type of features, such as the color, intensity, gradient orientation, and filter response.
			
			Since the number of pixels in each patch is small, it is preferable for the dimension $d$ to be low to ensure that the estimation of the covariance matrices of the patch Gaussians in the next step is robust. In this work, we extract eight-dimensional pixel features defined as:
			\vspace{-1mm}
			\begin{equation}
			\mathvc{f}_i = [ y, M_{0^\circ}, M_{90^\circ},  M_{180^\circ}, M_{270^\circ},  R, G, B ]^T, 
			\label{eq:pixelfeature}
			\end{equation}
			where $y$ is the pixel location in the vertical direction, $M_{\theta \in \{0^\circ, \dots, 270^\circ\} }$ are the magnitudes of the pixel intensity gradient along four orientations, and $R, G, B$ are the color channel values. Each dimension of \mathvc{f}$_i$ is linearly stretched to the range [0, 1] to equalize the scales of the different feature values.
			
			The pixel location is introduced to leverage the spatial information within each region. 
                        Using the vertical image location only originates from the analysis in ~\cite{Ma14}: the person images tend to be well aligned in the vertical direction whereas changes in the pose/viewpoint cause a large misalignment in the horizontal direction.
                        Note that it would be preferable to set $y_i$ from the top (or center) of the current region as in ~\cite{CarreiraCBS15}. However, each pixel belongs to multiple regions and such a setting would increase the computational complexity. Since person images are coarsely aligned, we directly set $y_i$ from the top of the image.
			
			The gradient information is introduced to provide the textural information of clothes. 
			The gradient orientation $O = {\rm arctan}({I_y}/{I_x})$ is calculated from the $x$- and $y$-derivatives $I_x, I_y$ of the intensity $I$. We quantize the orientation into four bins: $O_{\theta \in\{0^\circ,90^\circ,180^\circ,270^\circ\}}$. To complement the loss of information by the quantization, we use soft voting into two nearby orientation bins. The voting weights are linearly determined from the distances from the quantized orientations. We focus on high gradient edges by multiplying the gradient magnitude $M=$ {\footnotesize $\sqrt{I^2_y + I^2_y} $} by the quantized orientation ${O_{\theta}}$ to obtain the oriented gradient magnitude:  $M_{\theta} = MO_{\theta}$.
			
			Color information is the most important cue for person re-id. We use the color channel values of the most basic color space: RGB. Other color spaces, \eg Lab, HSV, and YCbCr, may be used. 
			In fact, we will extend our pixel features in different color spaces ($\S$\ref{sec:fusion}).
			
			\vspace{-2mm}
			\subsection{Patch Level Summarization}
			\label{sec:patch}
                        After extracting the pixel features inside a patch, we summarize them via the most classical parametric distribution, which has the mean and covariance as parameters: Gaussian distribution (Fig.~\ref{fig:flow} (b-ii)). For every patch $s$, we model the feature vectors as the patch Gaussian $\mathcal{N}(\mathvc{f};\mathvc{\mu}_s, \mathvc{\Sigma}_s )$  defined as,
			\begin{equation}
			\hspace{-1mm} \mathcal{N}(\mathvc{f};\mathvc{\mu}_s, \mathvc{\Sigma}_s ) =  \frac{\mathrm{exp} \left(- \frac{1}{2} (\mathvc{f} - \mathvc{\mu}_s)^T \mathvc{\Sigma}_s^{-1} (\mathvc{f} - \mathvc{\mu}_s) \right)}{ (2\pi)^{d/2} |\mathvc{\Sigma}_s| },
			\end{equation}
			where \mathvc{\mu}$_s$ is the mean vector and \mathvc{\Sigma}$_s$ is the covariance matrix of the sampled patch $s$. 
			The mean vector and the covariance matrix, respectively, are estimated by:
			$\mathvc{\mu}_s = \frac{1}{ n_s}  \sum_{ i \in \mathcal{L}_s }  \mathvc{f}_i $ and 
			$\mathvc{\Sigma}_s = \frac{1}{n_s-1}  \sum_{ i \in \mathcal{L}_s }  (\mathvc{f}_i - \mathvc{\mu}_s )(\mathvc{f}_i - \mathvc{\mu}_s)^T, $
			where $\mathcal{L}_s$ is the area of the sampled patch $s$ and $n_s$ denotes the number of pixels in $\mathcal{L}_s$.
			
			Note that the densely sampled mean vectors and covariance matrices can be efficiently calculated by using integral images~\cite{TuzelPM08}. Since regions can be overlapped, we construct the integral images of pixel features for an overall person image rather than creating them for each region. 
			
Gaussian Mixture Model (GMM) might be used for a more precise description of distributions. Because a local patch is expected to consist of a small number of colors/textures, 
we assume that the unimodal Gaussian is sufficient for describing the distribution of its pixel features.
			
			\vspace{1mm}
                        \noindent {\bf \itshape Gaussian Embedding} \\
\noindent As explained above, our descriptors are summarized representations of patch Gaussians inside a region. This summarization requires mathematical operations to obtain the mean or covariance of the Gaussian.
From the viewpoint of information geometry, the space of probability distributions is considered as a Riemannian manifold to which the Euclidean operation cannot be applied directly~\cite{Amari}.
The space of the SPD matrices is also considered as a Riemannian manifold and the LERM~\cite{ArsignyFPA06} provides a solid approach to map a point on the manifold to the Euclidean tangent space via a principal matrix logarithm.

To leverage the LERM, we embed the patch Gaussians in the SPD matrix in a manner similar to a previous work~\cite{li2013novel}.
From the analysis in the information geometry literature~\cite{lovric2000},  the space of $d$-dimensional multivariate Gaussians can be embedded into the $d+1$ dimensional SPD matrices space denoted by $Sym^{+}_{d+1}$. 
We represent the $d$-dimensional patch Gaussian $\mathcal{N}(\mathvc{f}; \mathvc{\mu}_s, \mathvc{\Sigma}_s)$ into $Sym^{+}_{d+1}$ as \mathvc{P}$_s$:
			\begin{equation}
			\hspace{-2.2mm}\mathcal{N}(\mathvc{f};\mathvc{\mu}_s, \mathvc{\Sigma}_s) \sim \mathvc{P}_s = |\mathvc{\Sigma}_s|^{-\frac{1}{d+1}} \left[ \begin{array}{cc}  \mathvc{\Sigma}_s+\mathvc{\mu}_s \mathvc{\mu}^T_s  &  \mathvc{\mu}_s  \\ \mathvc{\mu}^T_s & 1 \end{array} \right]. \label{eq:patchGauss}
			\end{equation}
The covariance matrix of the local patch often becomes singular due to the lack of a sufficient number of pixels within the patch. We overcome this problem by adding the identity matrix \mathvc{I}$_d$ to $\mathvc{\Sigma}_s$ with a small positive constant value $\epsilon_s$: $\mathvc{\Sigma}_s  \leftarrow \mathvc{\Sigma}_s + \epsilon_s \mathvc{I}_d$.
                      	
			Note that the scale normalization defined in $\S$\ref{sec:scalenorm} is inherently adopted for the Gaussian matrix \mathvc{P}$_s$. 
			This can be confirmed as follows:
                       	Let \mathvc{G}$_s$ be an SPD matrix excluding the scaling term of \mathvc{P}$_s$ 
			and let \mathvc{G}$_s (i,j)$ be its $(i,j)$-th block, \ie
			$ \mathvc{G}_s = \footnotesize{ \left[ \begin{array}{cc}  \mathvc{\Sigma}_s + \mathvc{\mu}_s \mathvc{\mu}_s^T  &  \mathvc{\mu}_s  \\ \mathvc{\mu}_s^T &  1  \end{array} \right] } = \footnotesize{ \left[ \begin{array}{rr}  \mathvc{G}_s(1,1)  &  \mathvc{G}_s(1,2)  \\ \mathvc{G}_s(2,1) &  \mathvc{G}_s(2,2) \end{array} \right] }$.
			The property of the submatrix~\cite{matrix} indicates that 
			$|\mathvc{G}_s| = |\mathvc{G}_s(2,2)| | \mathvc{G}_s(1,1) - \mathvc{G}_s(1,2)\mathvc{G}^{-1}_s(2,2) \mathvc{G}_s(2,1)| = |1| |\mathvc{\Sigma}_s + \mathvc{\mu}_s\mathvc{\mu}_s^T - \mathvc{\mu}_s ( 1^{-1} ) \mathvc{\mu}_s^T | = |\mathvc{\Sigma}_s|. $
                        Consequently, we have $\mathvc{P}_s = |\mathvc{G}_s|^{-\frac{1}{d+1}}\mathvc{G}_s =  \eta(\mathvc{G}_s)$.
                        
			
			In order to describe the region distribution in a Euclidean operation, we then map each of the patch Gaussians \mathvc{P}$_s$ into a tangent space via a principal matrix logarithm (Fig.~\ref{fig:flow} (b-iii)).
			
			We then store the upper triangular (or equivalent lower triangular) part of the mapped matrix as a vector because the matrix is symmetric. 
			By considering the off-diagonal entries as being counted twice during the norm computation~\cite{TuzelPM08}, the matrix of the patch Gaussian $\mathvc{P}_s$ becomes an $m=\frac{1}{2}(d+1)(d+2)$ dimensional vector $\mathvc{g}_s$, defined as,
			\vspace{-1mm}
                        \begin{equation}
			\mathvc{g}_s  = {\rm vec}({\rm log} \mathvc{P}_s) = [ {\rm diag}({\rm log}\mathvc{P}_s)^T  \quad  \sqrt{2} {\rm offdiag}({\rm log} \mathvc{P}_s)^T]^T, \label{eq:halfvec}
			\end{equation}
                        where ${\rm diag}(\cdot)$ and ${\rm offdiag}(\cdot)$ respectively represent the operator to reshape the diagonal elements and the upper-triangular (half) off-diagonal elements of a symmetric matrix into a vector form.
                        
\subsection{Alternative Summarization} 
\label{sec:zmg}
Because the dimensionality of the patch Gaussian vector grows quadratically {\itshape w.r.t.} the size of the row or column of the SPD matrix, a hierarchical use of this embedding drastically increases the dimensionality. 
It is desirable to retain the size of the SPD matrix as small as possible, even as small as one dimension. Thus, we develop an alternative embedding method.
We assume a Gaussian distribution of which mean vector is fixed to the zero vector \ie \mathvc{\mu}$_s = \mathvc{0} = (0,...,0)^T$.
The {\bf Z}{\itshape ero {\bf m}ean {\bf G}aussian} ({\bf ZmG}) distribution $\mathcal{N}(\mathvc{f}; \mathvc{0}, \mathvc{\Sigma}_s)$ is given by,

			\begin{equation}
			\hspace{-1mm} \mathcal{N}(\mathvc{f};\mathvc{0}, \mathvc{\Xi}_s ) =  \frac{\mathrm{exp} \left(- \frac{1}{2} \mathvc{f}^T \mathvc{\Xi}_s^{-1} \mathvc{f}  \right)}{ (2\pi)^{d/2} |\mathvc{\Xi}_s| },
			\end{equation}
			where the covariance matrix is estimated by $\mathvc{\Xi}_s = \frac{1}{n_s-1}  \sum_{ i \in \mathcal{L}_s }  \mathvc{f}_i \mathvc{f}_i ^T $.
			Note that the covariance matrix $\mathvc{\Xi}_s$ coincides with the raw (non-central) moment~\cite{Hong16} and is often referred to as the autocorrelation matrix~\cite{Fukunaga}. 
			
			The autocorrelation matrix naturally holds the mean information of the pixel features and almost coincides with the left upper diagonal block of the Gaussian matrix in Eq.(\ref{eq:patchGauss}). Namely, 
			\begin{equation}
			\mathvc{\Xi}_s = \frac{1}{n_s -1}  \sum_{ i \in \mathcal{L}_s } \mathvc{f}_i \mathvc{f}_i^T  =  \mathvc{\Sigma}_s + \frac{n_s}{n_s-1} \mathvc{\mu}_s\mathvc{\mu}_s^T \label{eq:ac1}.
			\end{equation}
			
			A further reduction of the parameters of the Gaussian distribution could be obtained by using the diagonal covariance assumption as in~\cite{Li16}. However, this assumption overly simplifies the pixel feature distribution and largely degrades re-id accuracies. 
			
			\vspace{1mm}
                        \noindent {\bf \itshape Zero-mean Gaussian Embedding} \\
			\noindent Using the same embedding as the Gaussian embedding, we can represent 
			$\mathcal{N}(\mathvc{f}; \mathvc{0}, \mathvc{\Xi}_s)$ into $Sym^{+}_{d+1}$ as \mathvc{D'}$_s$:
			\begin{equation}
			\mathcal{N}(\mathvc{f};\mathvc{0}, \mathvc{\Xi}_s) \sim \mathvc{D'}_s = |\mathvc{\Xi}_s|^{-\frac{1}{d+1}} \left[ \begin{array}{rr}  \mathvc{\Xi}_s &  \mathvc{0}  \\ \mathvc{0}^T &  1  \end{array} \right].
			\label{eq:patchzeroGauss}
			\end{equation}
			The autocorrelation matrix can be regularized as $\mathvc{\Xi}_s \leftarrow \mathvc{\Xi}_s + \epsilon_s \mathvc{I}_d$ to ensure that the matrix is an SPD matrix.
			
                        The eigendecomposition of the diagonal block matrix and the definition of the principal matrix logarithm are used to derive the matrix values on LE tangent space as follows: 
			\begin{equation}
			{\rm log}\mathvc{D'}_s = \left[ \begin{array}{cc}  {\rm log} \mathvc{\Xi}_s  - \frac{{\rm Tr}({\rm log}\mathvc{\Xi}_s)}{d+1}\mathvc{I}_{d}  &  \mathvc{0}  \\ \mathvc{0}^T &  - \frac{{\rm Tr}({\rm log}\mathvc{\Xi}_s)}{d+1}  \end{array} \right]  .
			\end{equation}
			A $\frac{1}{2}d(d+1)+1$ dimensional patch Gaussian vector may be obtained by taking only independent elements.
			
			Since we assumed that the mean vectors of the Gaussian are commonly zero, another natural choice to embed a ZmG distribution into an SPD matrix is to use the autocorrelation matrix \mathvc{\Xi}$_s$ directly as in the 2AvgP~\cite{CarreiraCBS15}. 
			However, as we will verify in $\S$\ref{sec:scale}, the scaling of the SPD matrix is important for HGDs.
			Based on an analogy to Gaussian embedding, which adopts scale normalization, we propose to represent the $d$-dimensional patch Gaussian $\mathcal{N}(\mathvc{f}; \mathvc{0}, \mathvc{\Xi}_s)$ into $Sym^{+}_{d}$ as the following \mathvc{D}$_s$:
			
			\begin{equation}
			\mathcal{N}(\mathvc{f};\mathvc{0}_s, \mathvc{\Xi}_s) \sim \mathvc{D}_s = |\mathvc{\Xi}_s|^{-\frac{1}{d}} \mathvc{\Xi}_s. \label{eq:patchAC}
			\end{equation}
			
Similarly to the case of the Gaussian matrix, we apply the principal matrix logarithm and half-vectorization to the scale-normalized matrix \mathvc{D}$_s$ and obtain an $m'=\frac{1}{2}d(d+1)$ dimensional vector $\mathvc{g}'_s = {\rm vec}({\rm log}\mathvc{D}_s)$.
			
			The embeddings \mathvc{D'}$_s$ and \mathvc{D}$_s$, which are very similar except for an additional dimension in $\mathvc{D'}_s$, were found to perform similarly. Since the size of $\mathvc{D}_s$ is more compact, we use $\mathvc{D}_s$ as the ZmG embedding.
			From the next subsection onward, we denote the flattened vector and its dimensionality as $\mathvc{g}$ and $m$, respectively, in both cases of the Gaussian and ZmG embeddings to keep the notation simple. 
\vspace{-3mm}
\subsection{Region Level Summarization} 
\label{sec:region}
As a result of the pose variation in person images, the positions of local parts vary in different observations. 
This leads us to summarize the local patches into an orderless representation. More specifically, we summarize the flattened patch Gaussians in the previous subsections into a region distribution (Fig.\ref{fig:flow} (b-iv)).
For this summarization, we also use a Gaussian distribution that not only has the ability to describe the covariance but also the mean.

Again, a GMM could be used to describe the distribution more precisely. However, matching among GMMs requires a complex algorithm, mainly to choose the components that need to be compared with one another~\cite{li2013novel}.
In addition, GMM requires time-consuming EM-based parameter estimation for every region. 
			
The use of a Gaussian distribution for summarization entails considering the spatial property of patches as follows.
A person image often contains background regions that significantly differ in places. We therefore suppress the effect of background regions by introducing a weight for each patch in a manner similar to that of weighted color histograms~\cite{bazzani}. 
In most cases, the person is centered in each image; thus, a higher value is assigned to the patches that are closer to the center of the y-axis of an image: $w_s = {\rm exp}(-(x_s - x_c)^2/ 2\sigma^2 )$, where $x_c = W/2$, $\sigma = W/4$. Here, $x_s$ denotes the $x$-coordinate of the center pixel of patch $s$ and $W$ is the image width. 
Then we define the weighted mean vector and covariance matrix as
			\begin{eqnarray}
			\mathvc{\mu^{\mathcal{G}}} &=& \frac{1}{\sum_{s\in \mathcal{G}} w_s} \sum_{s\in \mathcal{G}} w_s \mathvc{g}_s,  \\
			\mathvc{\Sigma^{\mathcal{G}}} &=& \frac{1}{\sum_{s\in \mathcal{G}} w_s} \sum_{s\in \mathcal{G}} w_s (\mathvc{g}_s -  \mathvc{\mu^{\mathcal{G}}} )( \mathvc{g}_s -  \mathvc{\mu^{\mathcal{G}}} )^T, 
			\end{eqnarray}
			where $\mathcal{G}$ is the region in which the patch Gaussians are summarized. Similarly, the weighted autocorrelation matrix is defined by
			\begin{equation}
			\hspace{-28mm} \mathvc{\Xi^{\mathcal{G}}} = \frac{1}{\sum_{s\in \mathcal{G}} w_s} \sum_{s\in \mathcal{G}} w_s  \mathvc{g}_s \mathvc{g}_s^T.
			\end{equation}
			Here we regularize the covariance and autocorrelation matrices \mathvc{\Sigma}$^\mathcal{G}$, \mathvc{\Xi}$^\mathcal{G}$ with the parameter $\epsilon^\mathcal{G}$, \eg $\mathvc{\Sigma}^\mathcal{G} \leftarrow \mathvc{\Sigma}^\mathcal{G} + \epsilon^\mathcal{G} \mathvc{I}_{m}$.
			Using the mean vector and covariance matrix, we represent the region as the region Gaussian $\mathcal{N}(\mathvc{g};\mathvc{\mu}^\mathcal{G}, \mathvc{\Sigma}^\mathcal{G})$ or ZmG $\mathcal{N}(\mathvc{g};\mathvc{0}, \mathvc{\Xi}^\mathcal{G})$.
	
                        For matching among region descriptors, it is convenient to map the region Gaussian into a Euclidean space where the most of the matching methods such as metric learning are designed on. 
                        For this purpose, we embed an $m$-dimensional region Gaussian into SPD matrices in the same manner as in Eq.(\ref{eq:patchGauss}) or Eq.(\ref{eq:patchAC}):
			$\mathcal{N}(\mathvc{g};\mathvc{\mu}^\mathcal{G}, \mathvc{\Sigma}^\mathcal{G}) \sim \mathvc{Q} \in Sym_{m+1}^{+}$ or $\mathcal{N}(\mathvc{g}; \mathvc{0}, \mathvc{\Xi}^\mathcal{G}) \sim \mathvc{R} \in Sym_{m}^{+}$. 
			We then map \{\mathvc{Q}, \mathvc{R}\} into LE tangent space by using the principal matrix logarithm and half-vectorize it to form an $r$-dimensional feature vector \mathvc{z} where $r = \{\frac{1}{2}(m+1)(m+2), \frac{1}{2}m(m+1)\}$, respectively, for \{\mathvc{Q}, \mathvc{R}\} (Fig.\ref{fig:flow} (b-v)).
			
By extracting the region Gaussian for each of $G$ regions, we obtain feature vectors $\{\mathvc{z}_g\}_{g=1}^G $. In order to maintain the spatial location of these vectors, we concatenate them and form a feature vector (Fig.\ref{fig:flow}(b-vi)). Then the feature representation of a person image is given by $\mathvc{z} = [\mathvc{z}^T_1, .., \mathvc{z}^T_G ]^T$.

			\subsection{Fusion Descriptors}
			\label{sec:fusion}
			We fuse the descriptors extracted from the common pipeline explained in the previous subsections to form the final descriptor. The fusion is simply performed by concatenating HGDs on different pixel features or embeddings. 
			
                        \vspace{1mm}
                        \noindent{\bf \bf \itshape Different Color Space Fusion} \\
			\noindent
It has been proven that descriptors extracted from different color spaces are complementary~\cite{YangYYLYL14}.
We extract additional color information in GOG or ZOZ by replacing the RGB channel values in the pixel feature in Eq.(\ref{eq:pixelfeature}) with the values of three alternative color spaces $\{$Lab, HSV, nRGB$\}$. Here the nRGB is the normalized color space (\eg nR = R/(R+G+B)). Since there is a redundancy in this space, we only use $\{$nR, nG$\}$.Thus, the dimensions of the nRnG color space are $(d', m',r') = (7,36,703)$ for GOG and $(d', m',r') = (7,28,403)$ for ZOZ, whereas the dimensions of each $\{$RGB, Lab, HSV$\}$ color space are $(d,m,r) = (8,45,1081)$ for GOG and $(d,m,r) = (8,36,666)$ for ZOZ. Therefore, the dimensions of the fusion descriptor of each \{GOG, ZOZ\} is 3 (color spaces) $\times$ $G$ (regions) $\times$ $r'$ (dim.)  + 1 (color space) $\times$ $G$ (regions) $\times$ $r$ (dim.). In the experiment, we use seven regions ($G=7$), \ie the dimensions of GOG and ZOZ become 27,626 and 16,828, respectively.

			\vspace{1mm}
                        \noindent{\bf \bf \itshape Different Hierarchical Embedding Fusion} \\
			\noindent We extract the two descriptors GOG and ZOZ by using the same embedding for both patch/region Gaussians. 
                        To make the feature representations robust, we concatenate both the GOG and ZOZ descriptors.
			We refer to our final fused descriptor as the HGDs, with the overall dimensions of 44,450 = 27,622 + 16,828.
			\vspace{-1mm}
			\section{Analysis of Scale Normalization}
			\label{sec:scale}
			In this section, we analyze the role of the scale normalization of the SPD matrix. 
			Through empirical observation, we found that the diagonal elements of the symmetric matrix tend to be largely emphasized in LE tangent space and that the scale normalization alleviates this bias. 
			
\begin{figure}[t]
				\begin{center}
					\begin{minipage}[b]{0.49 \linewidth }
						\centering\subfigure[Random image patches]{\includegraphics[width=0.88 \linewidth]{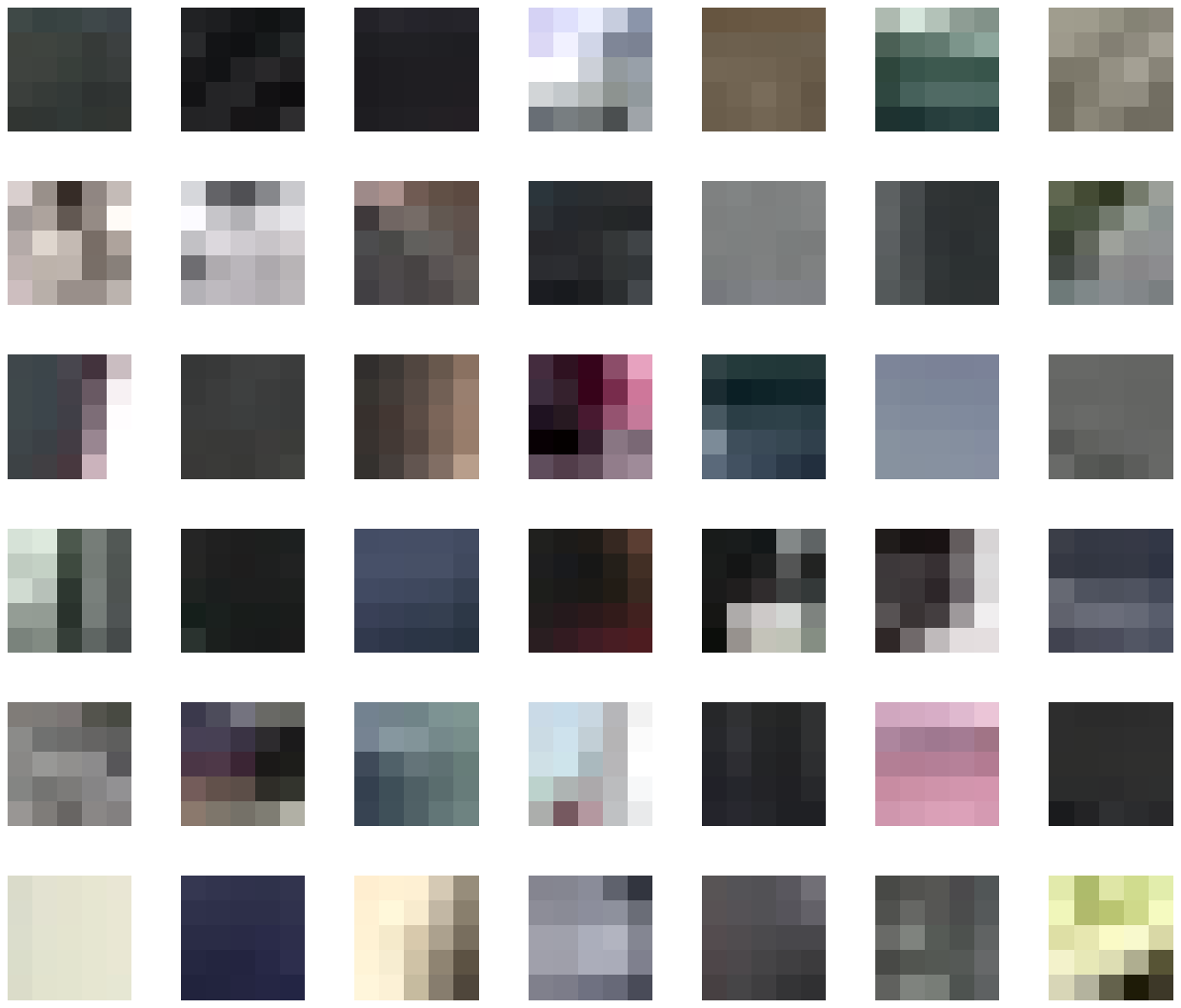}
							\label{fig:tangentspace}}
					\end{minipage}
					\begin{minipage}[b]{0.49 \linewidth }
						\centering\subfigure[Patch Gaussian matrices]{\includegraphics[width=0.95 \linewidth]{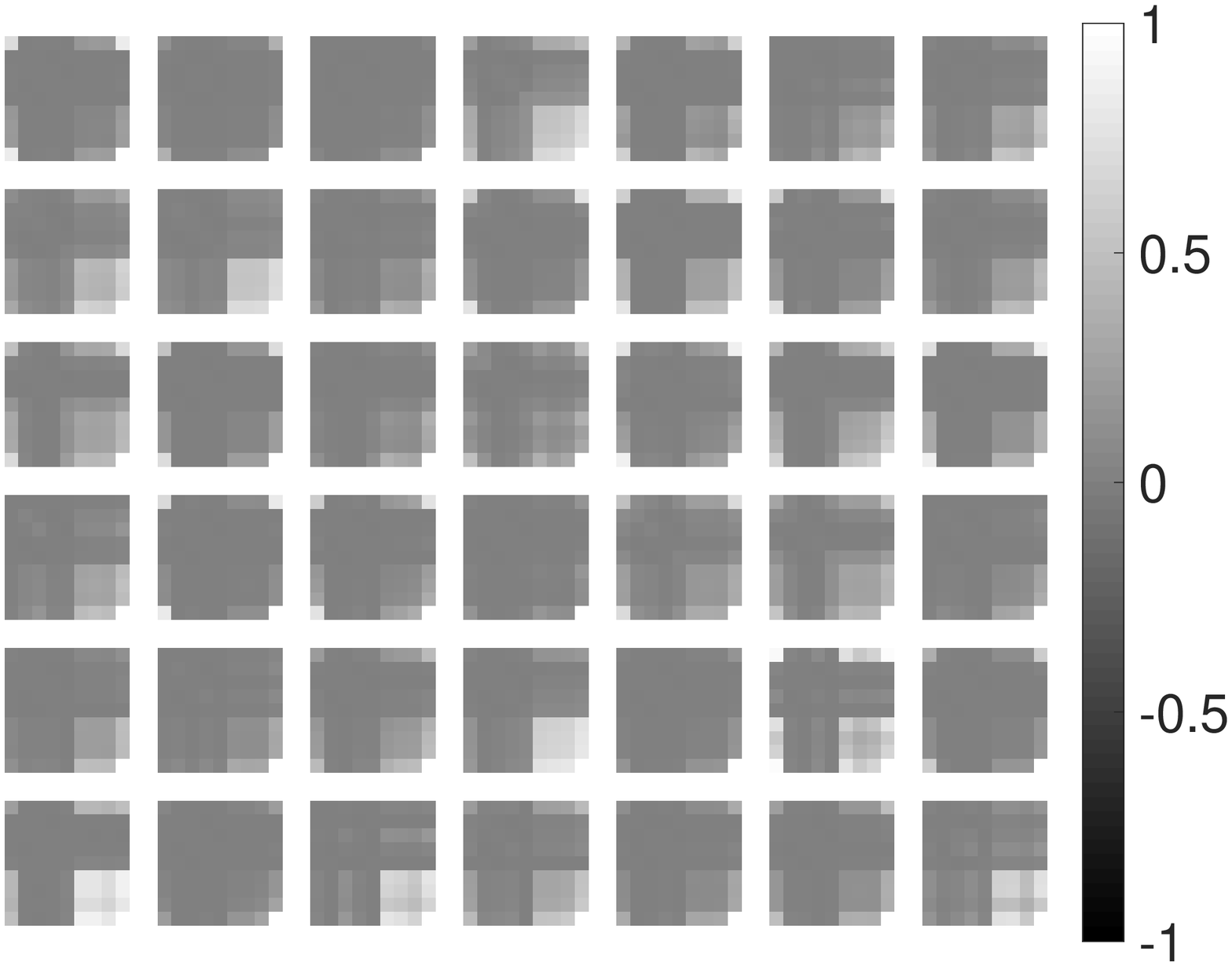}
							\label{fig:tangentspace}}
					\end{minipage}
                                        \vspace{-1mm} 
					\begin{minipage}[b]{0.49 \linewidth}
						\centering\subfigure[The matrices (b) on LE tangent space without the scale normalization]{\includegraphics[width=0.95 \linewidth]{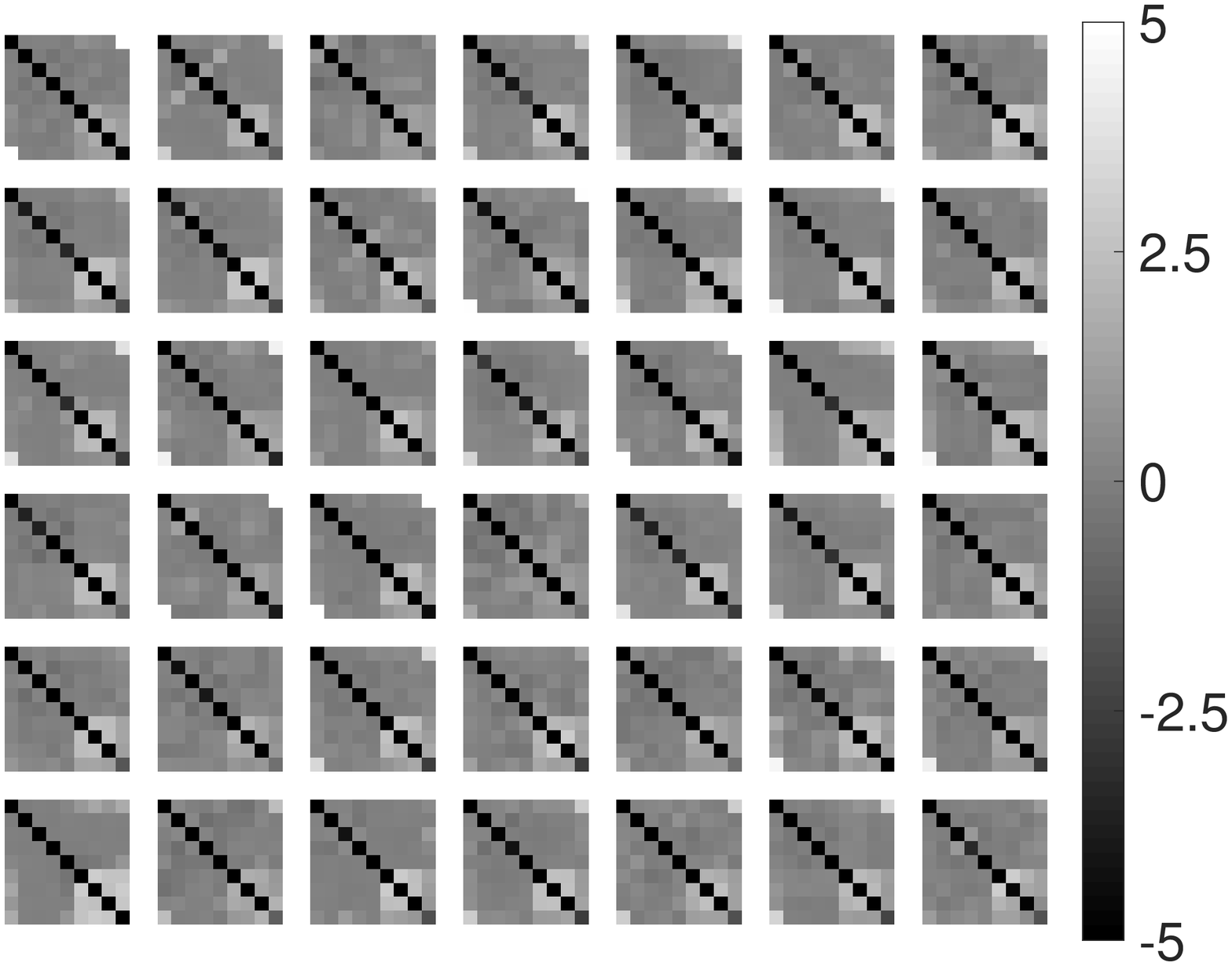}
							\vspace{-50mm}
							\label{fig:tangentspace}}
					\end{minipage}
					\begin{minipage}[b]{0.49 \linewidth}
						\centering \subfigure[The matrices (b) on LE tangent space with the scale normalization]{\includegraphics[width=0.95 \linewidth]{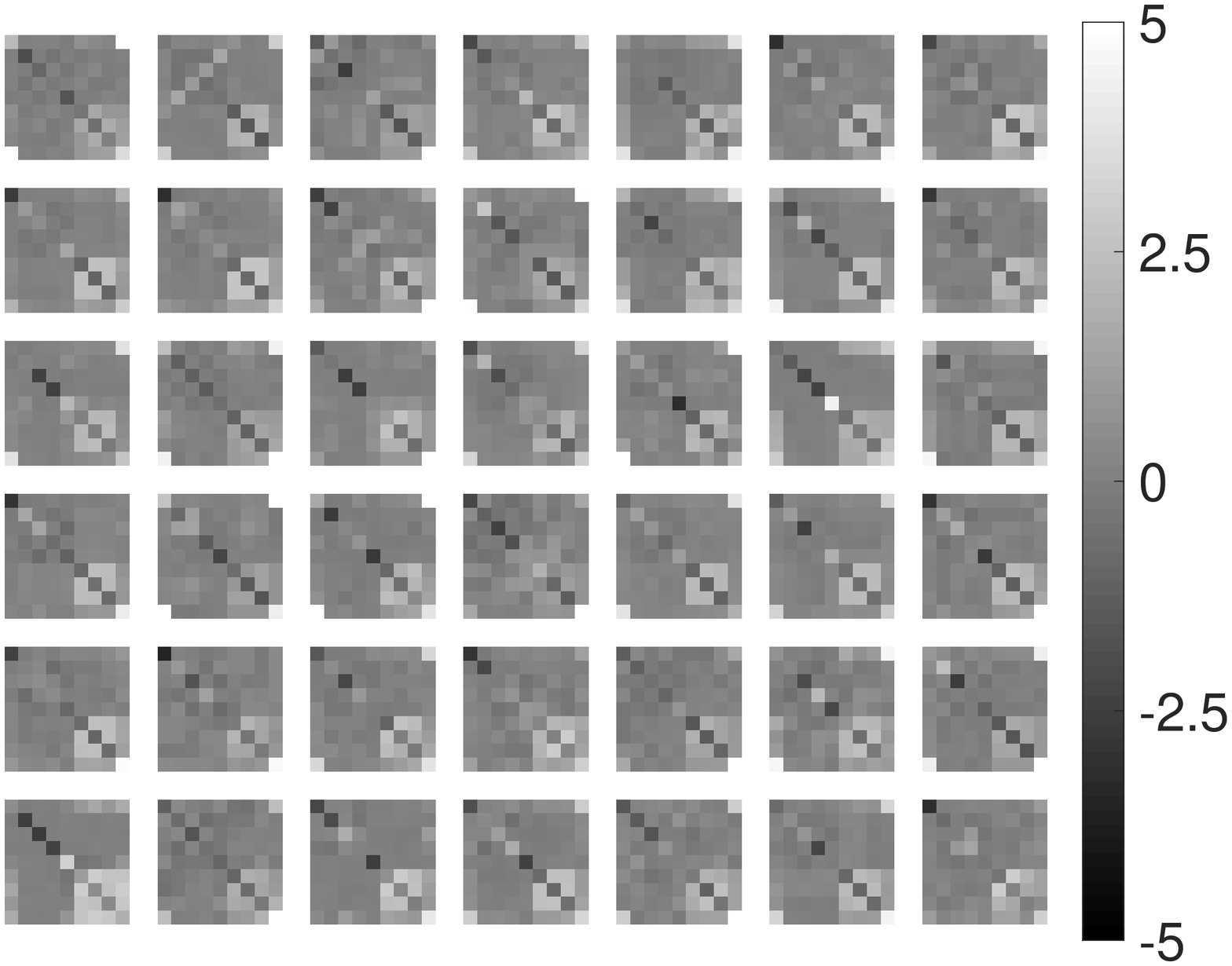}
							\label{fig:tangentspace}}
					\end{minipage}
					\vspace{-2mm}
				\end{center}
				\caption{Visualization of patch Gaussian matrices extracted from randomly sampled patches on the VIPeR dataset: (a) Image patches (5$\times$5 pixels). (b) Patch Gaussian matrices (8$\times$8 dims). (c,d) After obtaining the principal matrix logarithm of the (b) without/with the scale normalization.}
				\label{fig:log_matrix}
			\end{figure}

			We illustrate this effect by showing the patch Gaussian matrices on the VIPeR dataset~\cite{Gray} in Fig.~\ref{fig:log_matrix}\footnotemark. 
                        \footnotetext{ The parameter settings in the examples in this and the next sections are the same as those in the experimental section described in $\S$~\ref{sec:setup}.}
			We randomly sampled image patches from the database (Fig.~\ref{fig:log_matrix}(a)) and extracted their non-scaled patch Gaussian matrices \mathvc{G}$_s$ with the pixel features in Eq.(\ref{eq:pixelfeature}) (Fig.~\ref{fig:log_matrix}(b)). 
			By applying the principal matrix logarithm, we obtained the matrices in Fig.~\ref{fig:log_matrix}(c). 
			Fig.~\ref{fig:log_matrix} (d) shows the patch Gaussian matrices on LE tangent space when the scale normalization is applied to the matrices in Fig.~\ref{fig:log_matrix} (c). 
			It can be seen that the diagonal elements of the matrices commonly have large negative values in 
			Fig.~\ref{fig:log_matrix} (c) and they are alleviated in Fig.~\ref{fig:log_matrix}(d).
			\def\subfigcapskip{-5pt}  
			
                        \def\subfigcapskip{-3pt} 
			\begin{figure}[t]
				\begin{center}
					\begin{minipage}[b]{0.49 \linewidth }
						\centering\subfigure[Gauss]{ 
							\includegraphics[width=1 \linewidth]{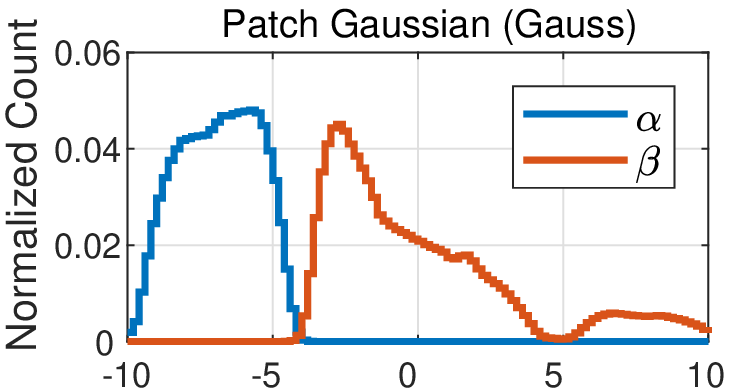}
						}
						\vspace{-5mm}
					\end{minipage}
					\begin{minipage}[b]{0.49 \linewidth }
						\centering\subfigure[ZmG]{ 
							\includegraphics[width=1 \linewidth]{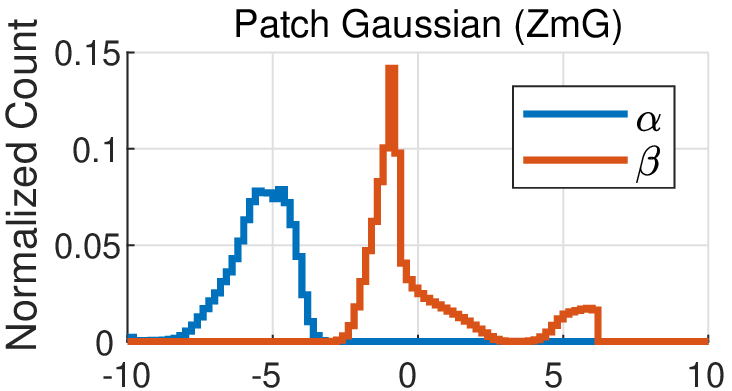}
						}
						\vspace{-5mm}
					\end{minipage}
					\vspace{-1mm}
				\end{center}
				\caption{ Histogram of decomposed log eigenvalues ($\alpha$ and $\beta_i, i=1,...,d+1$) on the VIPeR dataset: (a) Non-scaled patch Gaussian matrices \mathvc{G}. (b) Non-scaled autocorrelation matrices \mathvc{\Xi}. The count is normalized such that the sum of each histogram is one. }
				\label{fig:anal_eigen}
                                \vspace{-3mm}
			\end{figure}
%
			We first explain why the diagonal elements tend to be emphasized.
			Let $({\rm ln}\lambda_1, ..., {\rm ln} \lambda_{d+1})$ be the logarithm of the eigenvalues of a non-scaled patch Gaussian matrix $ \mathvc{G} \in Sym^{+}_{d+1}$.
			Let us decompose the values into the mean $\alpha = \frac{1}{d+1} (\sum_{j=1}^{d+1} \rm{ln}\lambda_j)$ and the residual $(\beta_1,...,\beta_{d+1})$. Using them, the $\rm{log}\mathvc{G}$ can be decomposed as
			\begin{eqnarray}
			{\rm log}\mathvc{G} &=& \mathvc{U} {\rm Diag} ( \beta_i + \alpha ) \mathvc{U}^T \nonumber \\
			&=& \mathvc{U} {\rm Diag} ( \beta_i ) \mathvc{U}^T + \alpha \mathvc{U}\mathvc{U}^T  = \mathvc{B} + \alpha \mathvc{I}_{d+1}. \:\:\:\:\:\: \label{eq:log2}
			\end{eqnarray}
			
			This decomposition implies that if the absolute value of the mean value is larger than the residual, \ie $|\alpha| \gg  |\beta_j|, j=1,..,d+1 $, then the diagonal elements of {\rm log}\mathvc{G} are expected to have larger absolute values than the off-diagonal elements.
			Actually, this situation often occurs in our HGDs. This is because local patches consist of fewer pixels. In such a case, only a few eigenvalues are expected to be large values, and most of them are expected to approximate zero.
			Since if $\lambda \rightarrow 0$ then ${\rm ln}{\lambda} \rightarrow -\infty $, the most logarithmic eigenvalues are expected to be large negative values.
			We show the histograms of $\alpha$ and $\beta$ on all images of the VIPeR dataset in Fig.~\ref{fig:anal_eigen}.
			
                        The HGDs construct the Gaussian/autocorrelation matrix upon the patch Gaussians with the {\itshape outer product} of the vectorized ${\rm log}\mathvc{G}$. Therefore, the large $\alpha$ dominates the region Gaussian. 
Recall the vectorization operation in Eq.(\ref{eq:halfvec}) and our concern with the case in which the diagonal elements of ${\rm log}\mathvc{G}$ are dominated by its diagonal elements, ${\rm diag}({\rm log}\mathvc{G}) = {\rm diag}(\mathvc{B} + \alpha\mathvc{I}_{d+1})={\rm diag}(\mathvc{B}) + \alpha \mathvc{1}_{d+1}\approx \alpha\mathvc{1}_{d+1}$ where $\mathvc{1}_{d+1} \in \mathbb{R}^{d+1} = (1,...,1)^T$. 
Therefore, $\mathvc{g} = {\rm vec}({\rm log} \mathvc{G}) = {\rm vec} ({\mathvc{B} + \alpha \mathvc{I}_{d+1}}) $ $= [ {\rm diag}(\mathvc{B} + \alpha \mathvc{I}_{d+1})^T  \quad  \sqrt{2} {\rm offdiag}(\mathvc{B})^T]^T  $
$\approx [ \alpha \mathvc{1}_{d+1}^T \; \mathvc{e}^T ]^T$
			where \mathvc{e} is a vector of which elements have relatively small absolute values compared to $\alpha$.
			Therefore, the patch Gaussian vectors in LE tangent space are expected to be dominated by $\alpha$ and thus the statistical values over the vectors are also expected to be dominated by it. 
			For example, the autocorrelation matrix (equivalent to the upper-left block of the Gaussian matrix) of \mathvc{g} becomes,
			\begin{equation}
			\hspace{-0.3mm} \mathvc{\Xi}^{\mathcal{G}} \approx  \frac{1}{\sum_{s\in \mathcal{G}}w_s} \sum_{s \in \mathcal{G}} w_s \left[ \begin{array}{rr}  \alpha^2_s \mathvc{1}_{d+1} \mathvc{1}_{d+1}^T  &  \alpha_s \mathvc{1}_{d+1}\mathvc{e}_s^T  \\  \alpha_s \mathvc{e}_s \mathvc{1}_{d+1}^T   &  \mathvc{e}_s\mathvc{e}_s^T  \end{array} \right]. 
			\end{equation}
			Since it is expected to be $\vert \alpha_s^2 \vert > \vert \alpha_s e_s \vert > \vert e_s e_s \vert$, where $e_s$ is an arbitrary component of \mathvc{e}$_s$, the region autocorrelation matrix is expected to be largely dominated by $\alpha_s^2$.
			This means only the mean logarithmic eigenvalue of the patch Gaussians is largely reflected in the region Gaussian.
			Therefore, removing the mean logarithmic eigenvalue from the patch Gaussian vector can improve the performance of HGDs.
			
                        \vspace{-1mm}
			\section{Norm Normalization of HGDs }
			\label{sec:normalization}
Norm normalization is important in order to equalize the ranges of the feature vectors and control the distance measure between them~\cite{Xie16}. In~\cite{SanchezPMV13}, it is pointed out that the norm normalization can help to improve the recognition accuracies for any of the high-dimensional features. Since the HGDs are high dimensional, we normalize the descriptor by using the L2 norm normalization, which is the most widely adopted form of normalization.

We observed that dimensions with common high/small values among different images exist within HGDs. 
Possible reasons are the following. There exist largely biased components of pixel features such as those for which the gradient magnitude is distributed sparsely in images, and the color intensity is distributed more uniformly. In addition, we assume that the SPD matrix has a common structure, \eg the last row or column in the Gaussian matrix is the mean vector in the Gaussian embedding. In such a case, the embedded SPD matrices will be similar apart from the identity matrix which is the origin of LE tangent space ($\S$~\ref{sec:LE}). 
In such a case, the cosine distance, \ie the Euclidean distance on L2 normalized features, would be dominated by the biased dimensions, and therefore decreases the discriminative ability.
	
			As a remedy for these biased dimensions, we investigate the use of two mean removal methods before L2 normalization.
			For the fusion descriptor, we normalize each of the HGDs extracted on four color spaces before concatenating them.
			
\begin{figure}[t]
				\begin{center}
                                        \includegraphics[width=1 \linewidth]{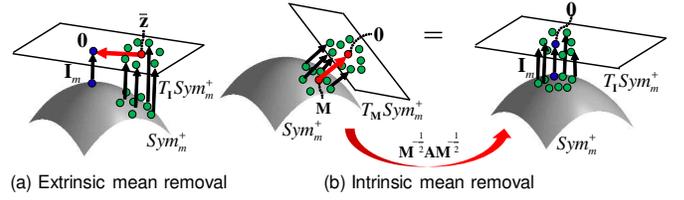}
					\vspace{-10mm}
				\end{center}
				\caption{ Two mean removal methods that we apply before feature norm normalization. (a) Extrinsic mean removal. (b) Intrinsic mean removal.
                                \label{fig:meanremoval}}
				\vspace{-2mm}
			\end{figure}

			\vspace{1mm}
                        \noindent{\bf \itshape Extrinsic Mean Removal + L2 Norm Normalization (E-L2) } \\
			\noindent\label{sec:normalization}In the first normalization, we simply remove the mean vector of the training samples directly on LE tangent space, of which calculation has a small computational cost (Fig.~\ref{fig:meanremoval} (a)). The normalization becomes the following:
			\begin{equation}
			\mathvc{\hat{z}} = \left(\mathvc{z} - \overline{\mathvc{z}} \right) / \Vert \mathvc{z} - \overline{\mathvc{z}} \Vert_2, 
			\end{equation}
where $\overline{\mathvc{z}}$ is the sample mean of the GOG or ZOZ. 
			
			A similar normalization was proposed for the Bag-of-Words representation to reflect co-missing words for cosine similarity~\cite{JegouC12}. In contrast, we employ the normalization to remedy the effect of biased dimensions on LE tangent space. 
			
			\vspace{1mm}
                        \noindent{\bf \itshape Intrinsic Mean Removal + L2 Norm Normalization (I-L2) } \\
			\noindent In the second normalization, we consider the intrinsic statistics of the Riemannian manifold~\cite{Pennec2006}.
			Here, we assume the region Gauss/ZmG matrices are not vectorized. If all matrices are similar apart from the identity matrix, the vector on the tangent space will be biased. 
			The natural choice to remove this bias is to use the tangent space of the mean point of training matrices. 
			
Let $\mathvc{A}$ be one of the region Gauss/ZmG matrices  $\{ \mathvc{Q}, \mathvc{R} \}$ and $m$ be the column (row) size of \mathvc{A}. 
Given a training set of the SPD matrices $\{ \mathvc{A}_i \in Sym_{m}^{+}\}_{i=1}^{N} $, the Riemannian center of mass \mathvc{M} is the point on $Sym_{m}^{+} $ that minimizes the sum of the squares of the Riemannian distance:
	 		\vspace{-1mm}
                        \begin{equation}
			\hat{\mathvc{M}} = \argmin_{\smathvc{M} \in Sym_{m}^{+} } \sum_{i=1}^{N} D_{\rm geo}^2 \left(\mathvc{A}_i, \mathvc{M}\right), 
                        \end{equation}
			where $D_{\rm geo}(\mathvc{A}, \mathvc{M})$ is the geodesic distance between \mathvc{M} and \mathvc{A}. 
			We use the AIRM distance of Eq.(\ref{eq:AIRM}). 
			By differentiating the error function with respect to \mathvc{M}, we see $\sum_{i=1}^{N} {\rm log}_{\smathvc{M}} \mathvc{A}_i = 0$, \ie \mathvc{M} must satisfy the mean tangent vectors on the pole to be zero. It is shown that the mean is unique and can be obtained by the following gradient descent procedure:
			\begin{equation}
                        \vspace{-0.5mm}
			\mathvc{M}^{t+1} = {\rm exp}_{\smathvc{M}^t} \left(  \frac{\delta}{N} \sum_{i=1}^{N} {\rm log}_{\smathvc{M}^t}  \mathvc{A}_i  \right), 
			\end{equation}
			where $\delta$ is the step size parameter and $t$ is the step number of iteration. This iteration is repeated until the computation converges.
			
			We use the mean matrix \mathvc{M} to map an SPD matrix \mathvc{A} into the tangent space by {\rm log}$_{\smathvc{M}}\mathvc{A}$. 
			By taking the orthogonal coordinates on the tangent space, the half vectorized representation is given by~\cite{Pennec2006}:
			\begin{equation}
                        \vspace{-0.5mm}
			\mathvc{z}' = \mathrm{vec}\left( {\rm log} \left(\mathvc{M}^{-\frac{1}{2}} \mathvc{A} \mathvc{M}^{-\frac{1}{2}} \right)\right).
			\end{equation}

			\def\subfigcapskip{-3pt} 
			\begin{figure}[t]
				\begin{center}
					\begin{minipage}[b]{0.49 \linewidth }
						\centering\subfigure[Standard L2 normalization]{ 
							\includegraphics[width=1 \linewidth]{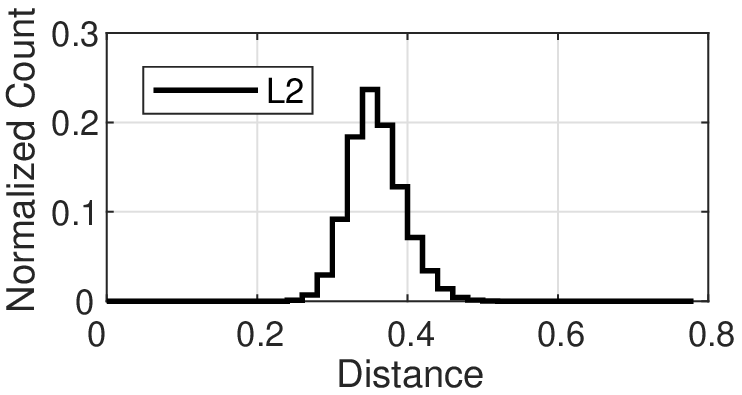}
						}
					\end{minipage}
					\begin{minipage}[b]{0.49 \linewidth }
						\centering\subfigure[Proposed L2 normalizations]{ 
							\includegraphics[width=1 \linewidth]{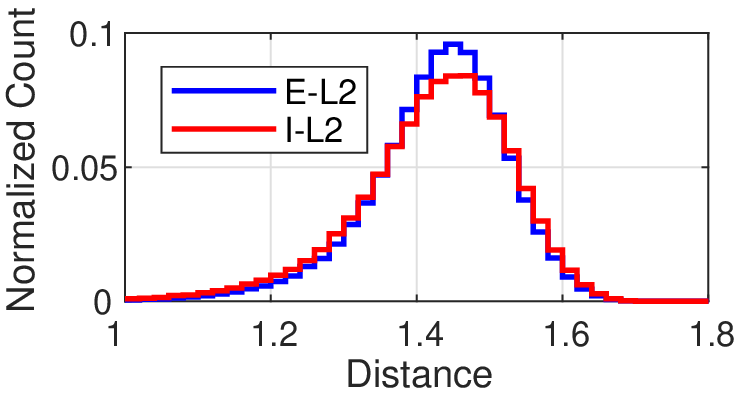}
						}
					\end{minipage}
					\vspace{-5mm}
				\end{center}
				\caption{ Histogram of distance on the VIPeR dataset: (a) Standard L2 normalization. (b) Proposed L2 normalizations. The count is normalized such that the sum of each histogram is one.  \label{fig:disthist}}
                               \vspace{-2mm}
			\end{figure}
			Note that the matrix $\mathvc{M}^{-\frac{1}{2}}$ is full rank because \mathvc{M} is an SPD matrix 
			and thus the transformed matrix $\mathvc{M}^{-\frac{1}{2}} \mathvc{A} \mathvc{M}^{-\frac{1}{2}}$ is also an SPD matrix.
			Therefore, we can interpret the space of $\mathvc{z}'$ as being LE tangent space of transformed matrices (Fig.\ref{fig:meanremoval}(b)).
                        Consequently, under the LERM, we can interpret the Euclidean distance on the tangent space of the pole as being the geodesic of the transformed matrices.
                        
                        We independently estimate the Riemannian mean for each region of the horizontal stripes. We map the region Gaussian matrices to the tangent space of each region and apply half-vectorization. Since in the tangent space, the mean vector on the training data is zero, we directly perform the normalization as \mathvc{\hat{z}'} = \mathvc{z'}$/\Vert\mathvc{z}'\Vert_2$.
                        
In Fig.~\ref{fig:disthist}, we compare the distance distribution on all possible image pairs of the VIPeR dataset when the proposed and standard normalizations are applied. In this comparison, we used the GOG descriptor with RGB pixel features.
We can confirm that by applying mean removal before the L2 normalization (Fig.~\ref{fig:disthist}(b)), the range of distance distributions becomes wider than the standard L2 normalization (Fig.~\ref{fig:disthist}(a)), both in the case of extrinsic and intrinsic mean removals. In addition, intrinsic mean removal has a slightly wider range of distance distribution than extrinsic mean removal.
This is because the mapping the data into the tangent space at the mean point similarly works as a whitening operation~\cite{barachant2013}.
\vspace{-3mm}
\section{Experiments}
\vspace{-0.5mm}
\subsection{Experimental Settings} 
\label{sec:setup}
\begin{figure}[t]
				\begin{center}
					\includegraphics[width=1 \linewidth]{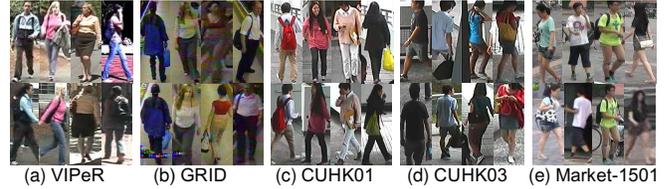}
					\vspace{-8mm} 
					\caption{Example images from the person re-id datasets.
						For each dataset, images in the same column represent the same person. }
					\label{fig:Datasets}
					\vspace{-4mm} 
				\end{center}
			\end{figure}

\vspace{-0.5mm}
\noindent{\bf Datasets and evaluation protocol.}
We use five benchmark datasets to evaluate our method: VIPeR~\cite{Gray}, GRID~\cite{LoyXG10}, CUHK01~\cite{LiZW12}, CUHK03~\cite{LiZXW14}, and Market-1501~\cite{zheng2015}. 
Example images of each of these datasets are shown in Fig.~\ref{fig:Datasets}.
All of these datasets are challenging because the images contain large variations in terms of their viewpoints, pose, illumination, occlusion, and background clutter. 
We evaluate the performance by the Cumulative Matching Characteristic (CMC) curves which is an expectation of finding the correct person in the top $r$ matches~\cite{Gray}.
For a measure to evaluate the whole CMC curves, we report the Proportion of Uncertainty Removed (PUR) which represents the uncertainly reduction by a given algorithm from the random ranking~\cite{PedagadiOVB13}.
For the Market-1501 dataset, we report the mean Average Precision (mAP) which considers both the prevision and recall of the retrieval process~\cite{zheng2015} since the gallery images contain multiple images for one person.

The {\bf \itshape VIPeR} contains 1,264 images with 632 persons captured in disjoint camera views. 
The {\bf \itshape GRID} dataset contains 1,275 images with 250 annotated persons and an additional 775 gallery images that do not belong to the annotated persons. 
Both the VIPeR and GRID datasets contain one image of each person with one camera view, thus we evaluate the performance with {\itshape single-shot} matching. 
We report an average of 10 random training/test splits, for each split only one half of people can be contained. 
The {\bf \itshape CUHK01} dataset contains 3,884 images of 971 persons. There are two images of each person in each camera view. Thus, as default, we carry out the evaluation with {\itshape multi-shot} matching, in which we calculate the distances between two persons by averaging the corresponding cross-view image pairs. We report the average of 10 random 485/486 person splits for the training/test sets.
The {\bf \itshape CUHK03} dataset contains 13,164 images of 1,360 persons with an average of 4.8 images of each person in each view. As default, we use the images that are {\itshape automatically detected} by the person detector, and evaluate the performance with {\itshape multi-shot} matching. We report the average result of 20 random 1,260/100 person splits for the training/test sets.
The {\bf \itshape Market-1501} dataset contains 32,668 bounding boxes of 1,501 persons. Each person is captured by six cameras at most, and two cameras at least. During testing, for each person, one query image in each camera is selected. We use a fixed 750/751 person split for the training/test set. As default, we report the results of the {\itshape single-query} evaluation on 3,386 query images.
\noindent{\bf Parameters of HGDs.} We resize each image in the datasets to 128$\times$48 pixels to facilitate evaluation with the common parameters of the descriptor. We extract the descriptor from seven overlapping horizontal strips ($G=7$). 
Each of the strips consists of 32$\times$48 pixels. By considering the trade-off between the computational time and the predictive accuracy, we extract local patches at two-pixel intervals ($p=2$) in each region. We set the local patch size to $5\times5$ pixels ($k=5$).
We set the regularization parameter for region Gaussian as $\epsilon^\mathcal{G} = \epsilon_{0} \rm{Tr}(\Sigma^\mathcal{G})$. 
In several patch Gaussians, the trace norm of the covariance matrix becomes nearly zero when the patch contains only nearly equal pixel values. Thus, we assign a small constant value to $\epsilon_s$ for the patch Gaussian, in which case we have $\epsilon_s = \epsilon_0 \rm{max}( \rm{Tr}(\Sigma_s), 10^{-2})$. We set $\epsilon_0 = 10^{-3}$ for both the patch and region Gaussians.

As default, we use a fused descriptor of four-color space and apply I-L2 normalization. For the normalization, we initialize the mean matrix with the log-Euclidean mean and set $\delta = 0.5$. The update of the mean matrix typically converges within 20 iterations. 

\noindent{\bf Distance Metrics.}
We evaluate the proposed descriptors by learning the distance metric with Cross-view Quadratic Discriminant Analysis (XQDA)~\cite{Liao15}. 
The XQDA learns a discriminative subspace and a distance metric simultaneously, and is able to select the optimal dimensionality automatically.

			\def\subfigcapskip{0pt}  
\vspace{-1mm} 
\subsection{Component-wise Evaluation}
We evaluate the components of our descriptors in terms of the following aspects: 
(1) Impact of the hierarchical Gaussian embedding; (2) Impact of the scale normalization; (3) Impact of the norm normalization; (4) Running time.
			
\noindent{\bf Impact of the hierarchical Gaussian embedding.}
We compare different embedding methods for feature summarization including both global and hierarchical methods.  
For global distribution embeddings of pixel features inside each region, we carry out a comparison with the mean vector (Mean), covariance matrix (Cov), Zero mean Gaussian (ZmG), and Gaussian (Gauss). For a fair comparison, we commonly adopt the weighted embedding for all descriptors, defined as follows:
			\begin{itemize}
                        	\vspace{-0.5mm}
				\setlength{\leftskip}{-0.5cm}
				\item Mean: $\mathvc{\mu}' = \frac{1}{\sum_{i \in \mathcal{G}} w_i } \sum_{i \in \mathcal{G} } w_i \mathvc{f}_i, $
				
				
				\item Cov: $\mathvc{\Sigma}' = \frac{1}{\sum_{i \in \mathcal{G}} w_i }  \sum_{ i \in \mathcal{G} } w_i ( \mathvc{f}_i - \mathvc{\mu}') ( \mathvc{f}_i - \mathvc{\mu}') ^T, $
				
				\item ZmG: $\mathvc{\Xi}' = \frac{1}{\sum_{i \in \mathcal{G}} w_i }  \sum_{ i \in \mathcal{G} } w_i  \mathvc{f}_i \mathvc{f}_i^T, $
				
				\item Gauss: $ \mathvc{P}' = \left[ \begin{array}{cc}  \mathvc{\Sigma}' + \mathvc{\mu}' \mathvc{\mu}'^T  &  \mathvc{\mu}'  \\ \mathvc{\mu}'^T &  1  \end{array} \right],$ \\
                                \vspace{-1mm}
			\end{itemize}
			here the pixel weight $w_i$ is determined in the same manner as $w_s$.

For two-level embeddings, we compare the performance with Covariance-Of-Covariance (COC) embedding, which is defined as follows:
\begin{itemize} 	        \vspace{-0.2mm}
				\setlength{\leftskip}{-0.5cm}
				\item COC: $\mathvc{S} = \frac{1}{\sum_{s\in \mathcal{G}} w_s }  \sum_{s\in \mathcal{G}} w_s (\mathvc{h}_s -  \mathvc{\nu}  )( \mathvc{h}_s -  \mathvc{\nu} )^T $ 
				where \mathvc{h}$_s = \rm{vec}(\rm{log}(\mathvc{\Sigma}_s)) $ and $\mathvc{\nu} = \frac{1}{\sum_{s \in \mathcal{G}} w_s } \sum_{s \in \mathcal{G} } w_s \mathvc{h}_s $.
                                \vspace{-0.2mm}
\end{itemize}

We apply the LE tangent space mapping and half vectorization for all descriptors except Mean. In the same manner as HGDs, we regularize the covariance/autocorrelation matrices and concatenate the feature vectors of seven regions.

\begin{figure}[t]
				\begin{center}
					\begin{minipage}[b]{1 \linewidth }
						\centering\subfigure[RGB]{ 
							\includegraphics[width=0.32 \linewidth]{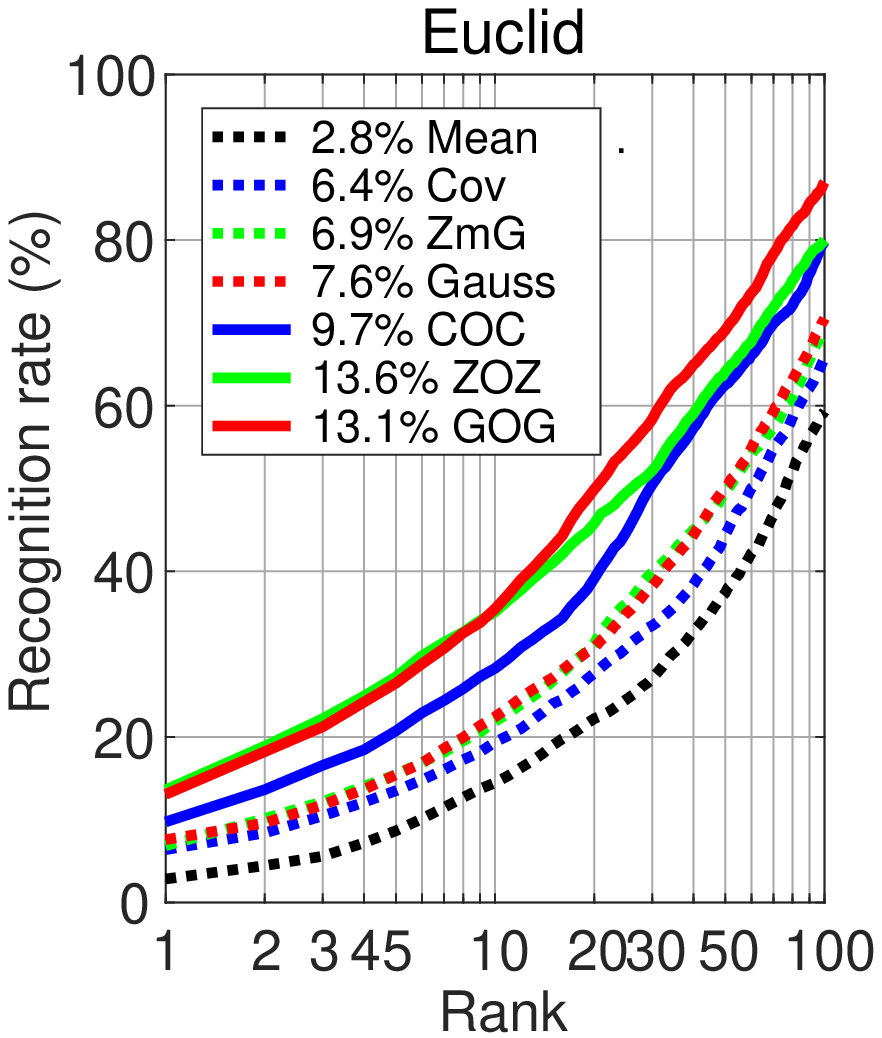}
							\includegraphics[width=0.32 \linewidth]{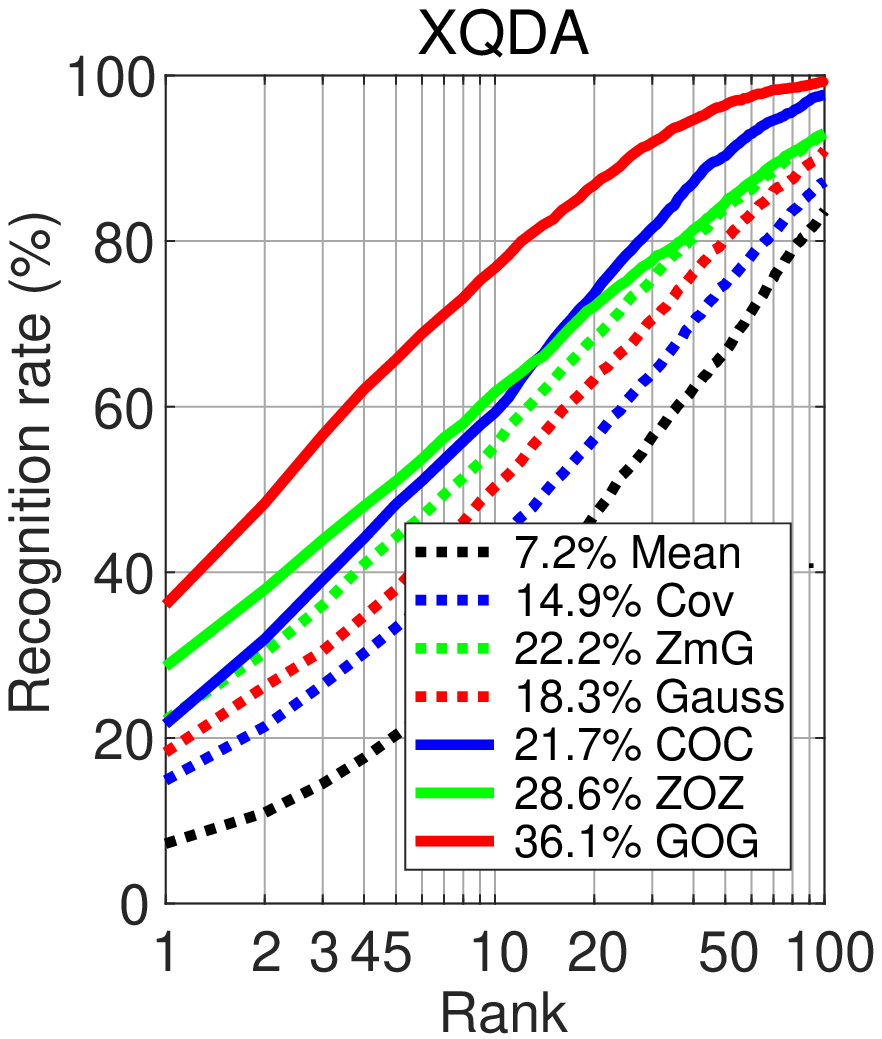}
							\includegraphics[width=0.32 \linewidth]{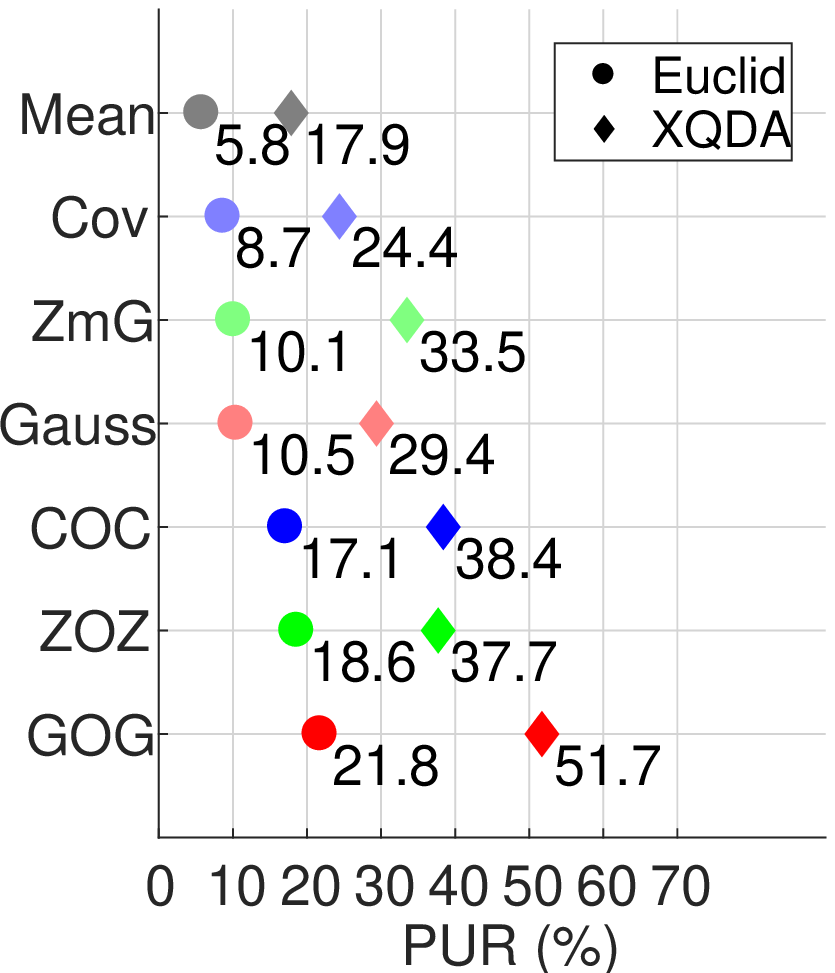}
						}
					\end{minipage}
					\begin{minipage}[b]{1 \linewidth }
						\centering\subfigure[Color space fusion]{ 
							\includegraphics[width=0.32 \linewidth]{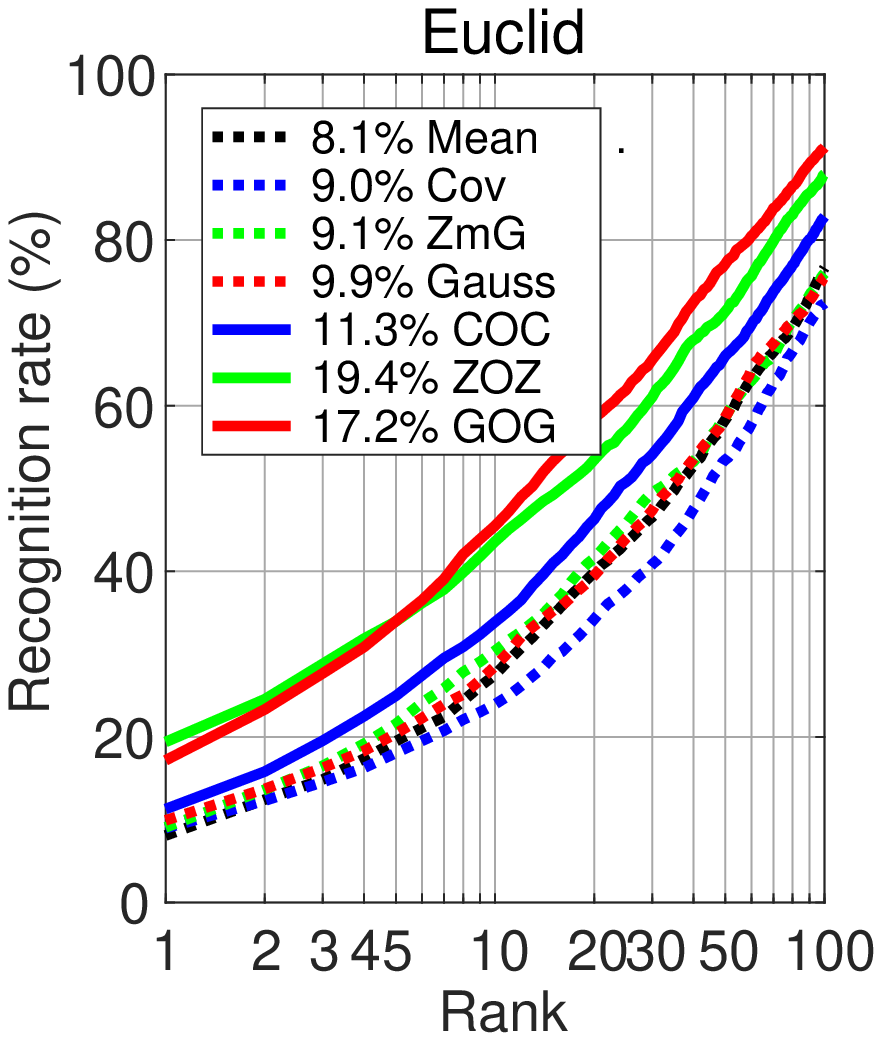}
							\includegraphics[width=0.32 \linewidth]{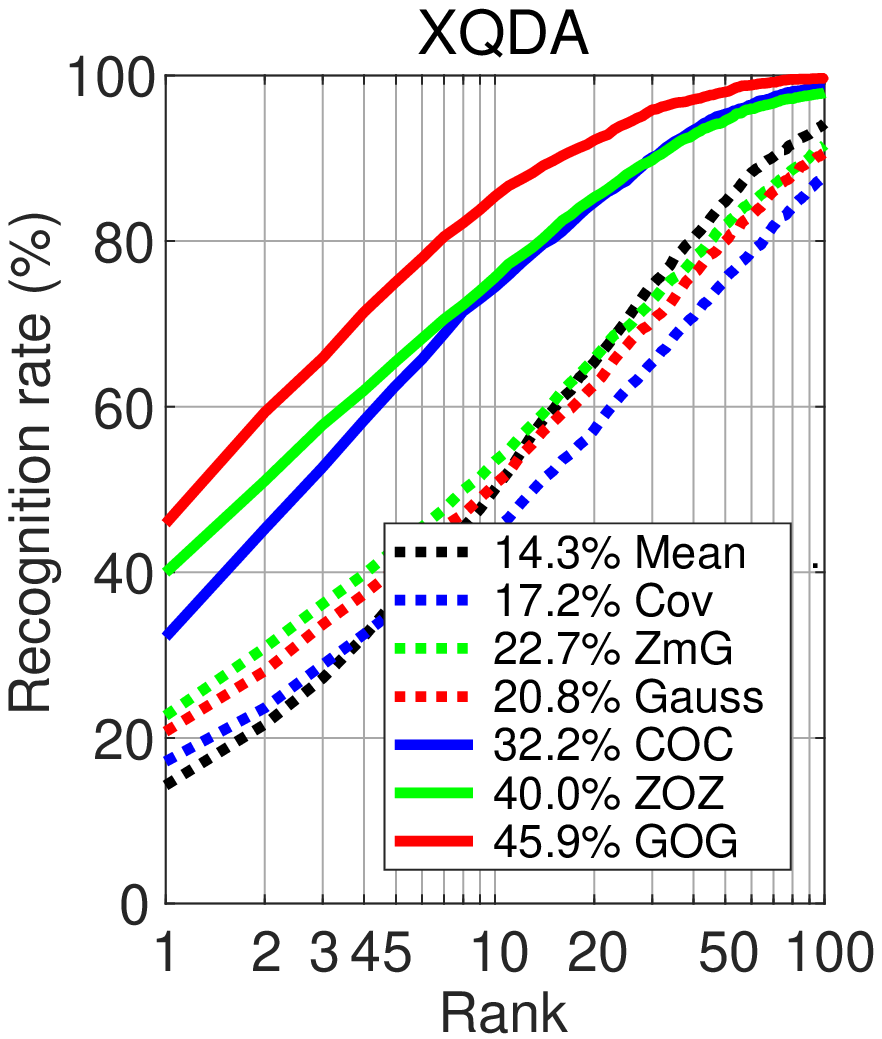}
							\includegraphics[width=0.32 \linewidth]{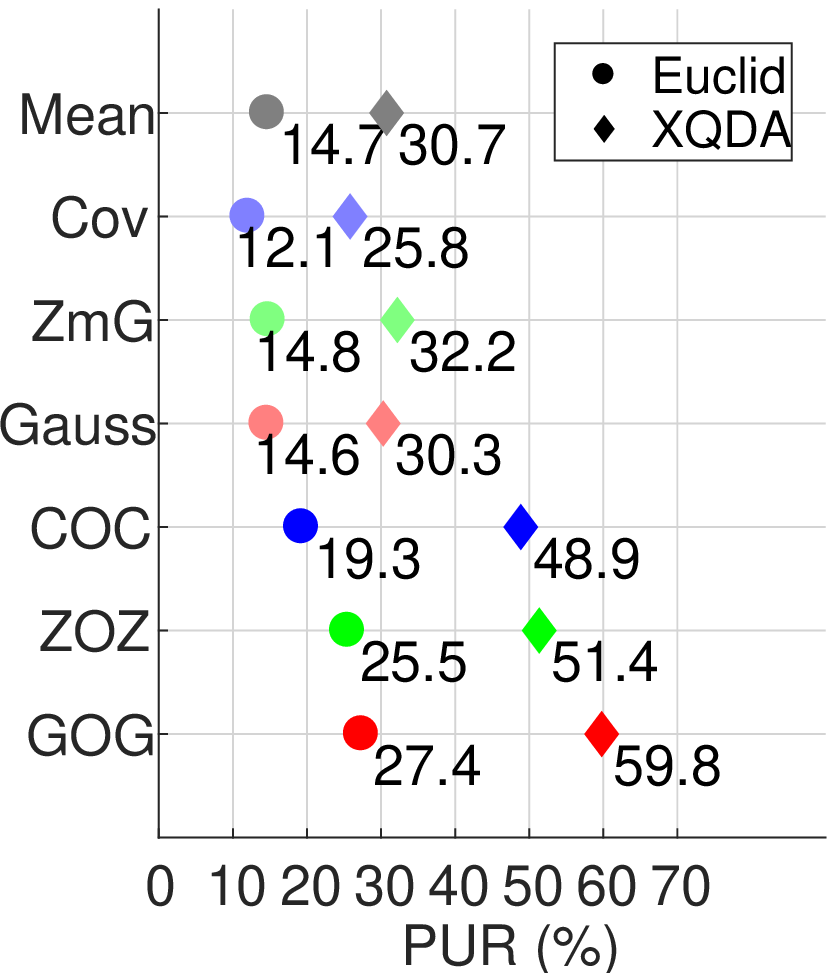}
						}
					\end{minipage}
					\begin{minipage}[b]{1 \linewidth }
						\centering\subfigure[Color space fusion + scale normalization]{ 
							\includegraphics[width=0.32 \linewidth]{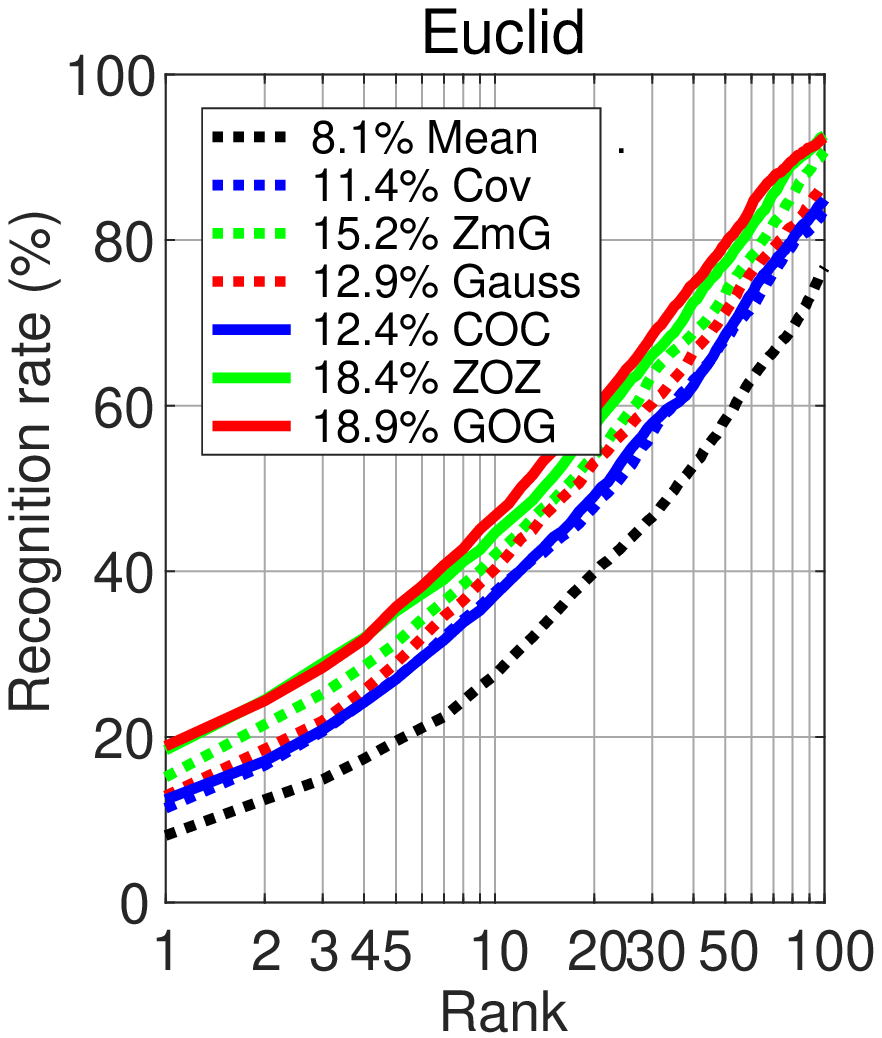}
							\includegraphics[width=0.32 \linewidth]{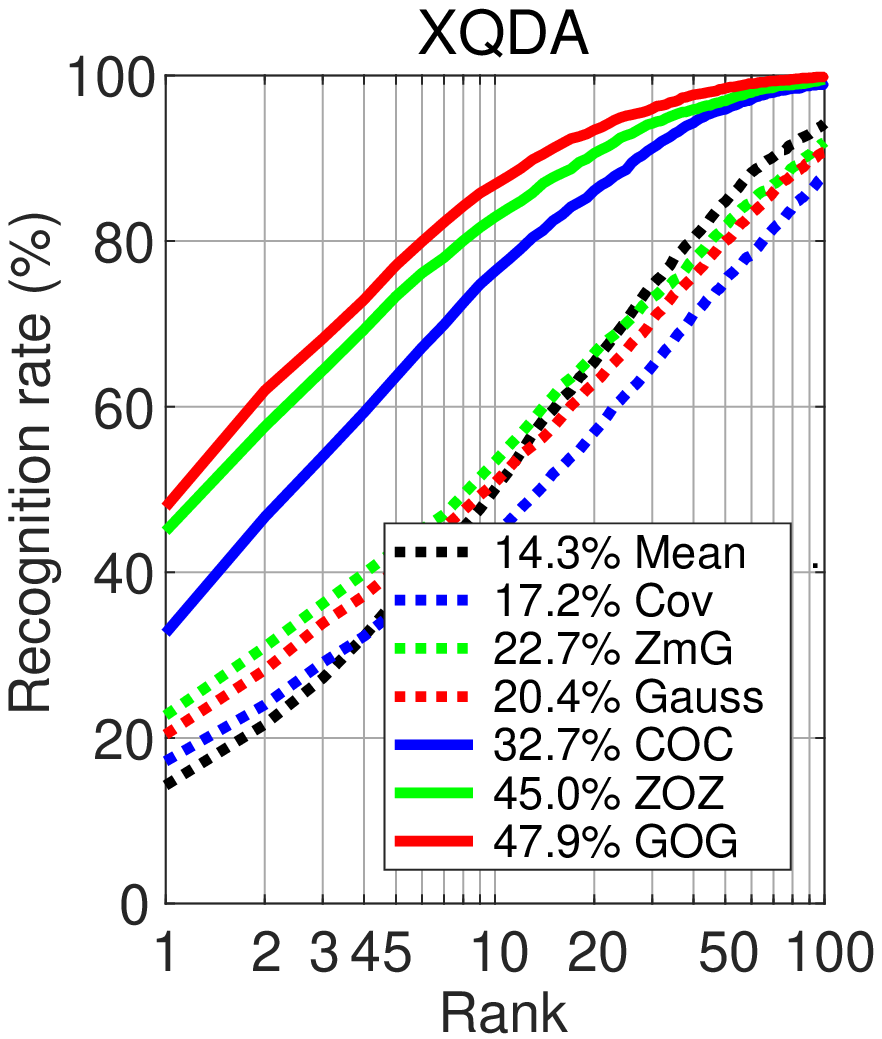}
							\includegraphics[width=0.32 \linewidth]{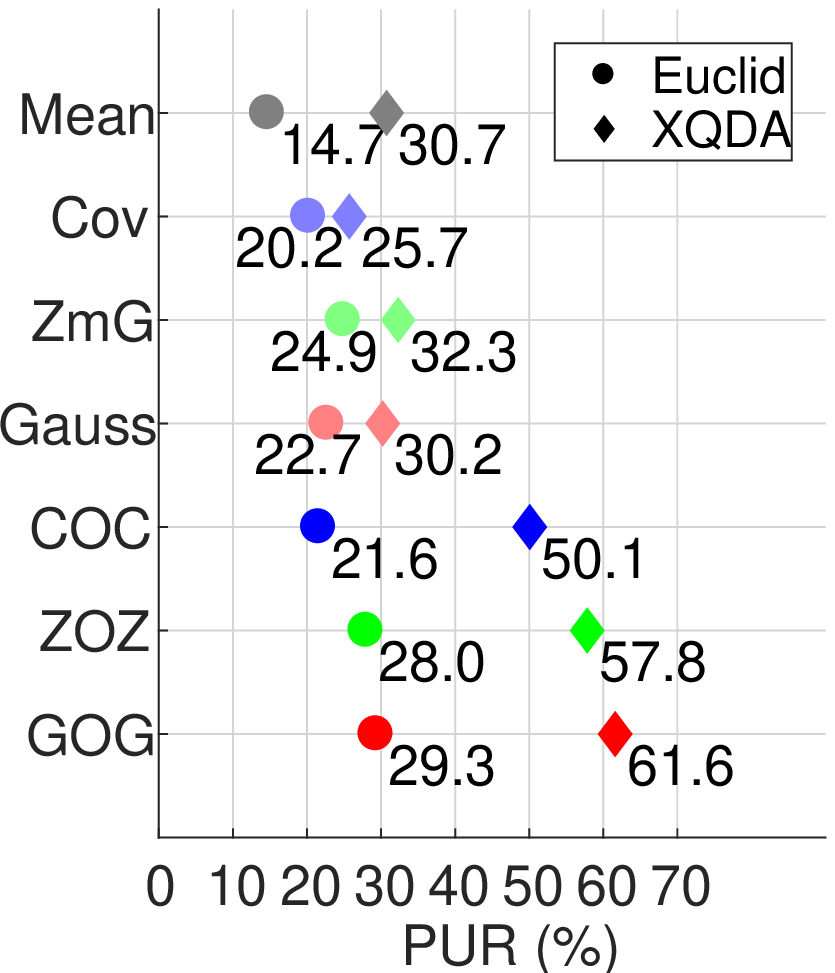}
						}
					\end{minipage}
					\begin{minipage}[b]{1 \linewidth }
						\centering\subfigure[Color space fusion + scale + norm normalizations]{ 
							\includegraphics[width=0.32 \linewidth]{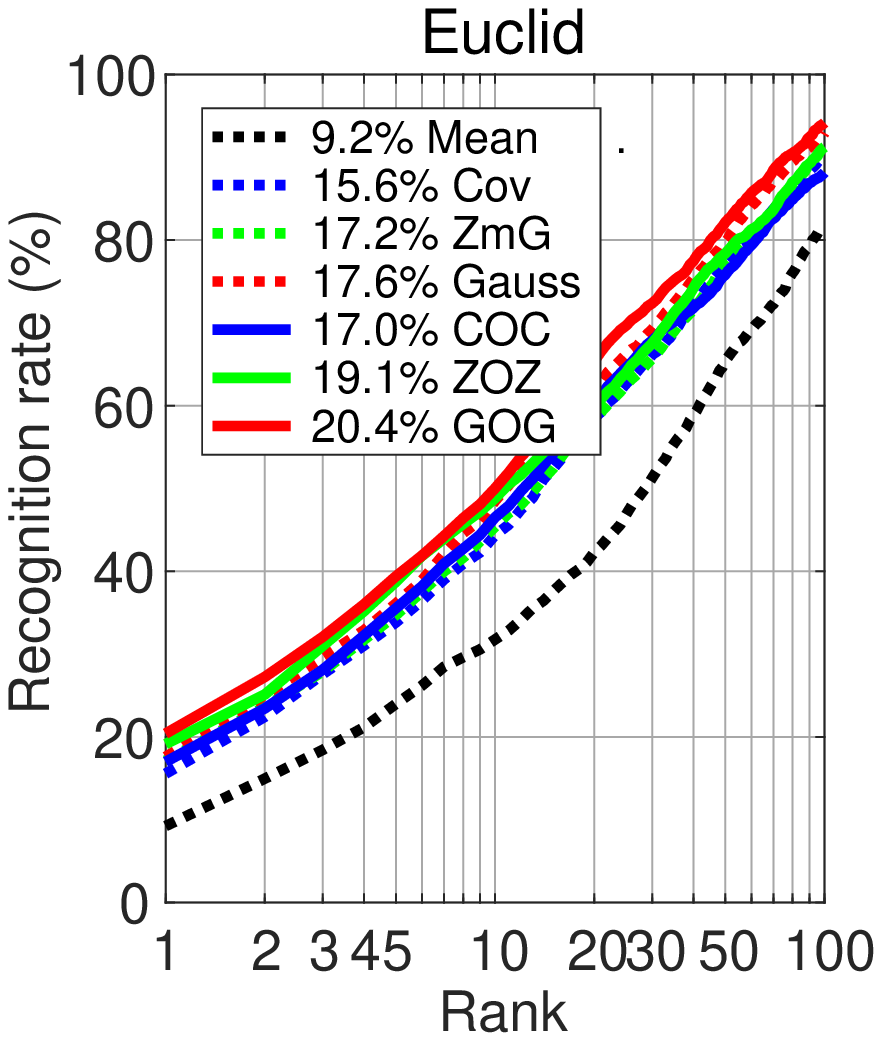}
							\includegraphics[width=0.32 \linewidth]{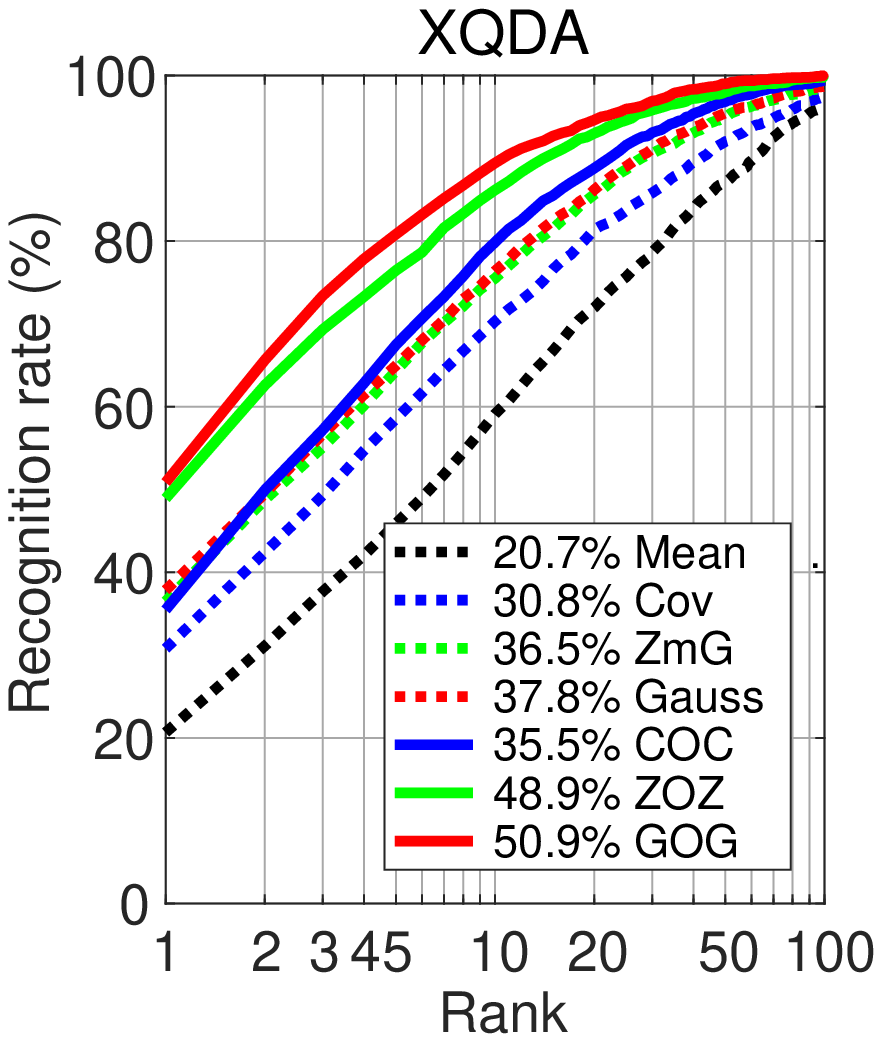}
							\includegraphics[width=0.32 \linewidth]{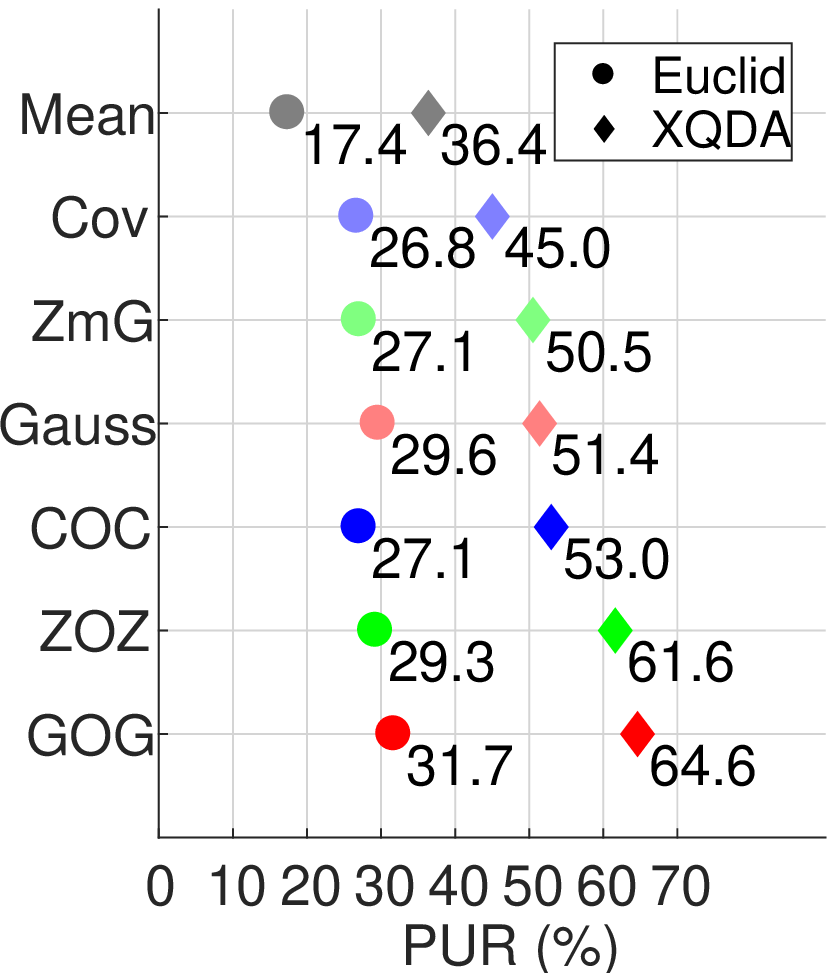}
						}
					\end{minipage}
                                        \vspace{-3mm}
					\caption{ Feature embedding analysis on the VIPeR dataset: (a) Pixel features on RGB color space. (b) Color space fusion without the scale and norm normalizations. (c) Color space fusion with the scale normalization. (d) Color space fusion with the scale and I-L2 norm normalizations. The numbers on the CMC curves indicate the rank-1 rates. }
					\label{fig:performance}
					\vspace{-6.5mm}
					\label{fig:embedding}
				\end{center}
			\end{figure}

In addition to the XQDA distance metric, we evaluate the performance with Euclidean distance to see the basic property of each embedding. We start the evaluation with the pixel features on RGB color space without the scale and norm normalizations. We then incrementally add the color space fusion and the two normalizations. The results on the VIPeR dataset are shown in Fig.~\ref{fig:embedding}. We show the results on the five datasets with the final model in Table~\ref{table:impact_hie} to show the generality of the comparison.

			\def\subfigcapskip{-2pt}  
From the results, we can see that:
(1) The performance trend is similar between the Euclidean and XQDA metrics: hierarchical and Gaussian embedding (GOG) performs the best in all cases. This indicates that this embedding method is comparatively discriminative and robust to summarize pixel features.
(2) The hierarchical embeddings achieve higher identification scores than global embeddings when the same base embedding is used. For example, COC significantly outperforms Cov. These results confirm one of our motivations, \ie the hierarchical distribution is more discriminative than the global distribution.
(3) Gaussian embeddings are more effective than covariance embedding. Among the global embeddings, ZmG and Gauss outperform Cov and Mean. This confirms another motivation, \ie the need to use both the mean and covariance information of pixel features. Among the hierarchical embeddings, Gaussian embeddings (ZOZ and GOG) achieves higher scores than the covariance embedding (COC). This confirms the importance of using both the mean and covariance information in the region level embedding.
(4) Different color fusion improves the performance. Fig.~\ref{fig:embedding}(a) and (b) together indicate that color space fusion is more accurate than using only RGB color space in all descriptors.
(5) The scale and norm normalizations generally improve the performance. Fig.~\ref{fig:embedding}(b) and (c) together show that the scale normalization considerably improves ZOZ embedding of which performance becomes close to GOG embedding.
			\def\subfigcapskip{-2pt} 
			
			
			\begin{figure}[t]
				\begin{center}
					\begin{minipage}[b]{1 \linewidth }
						\centering\subfigure[ Scale normalization (left: Euclidean distance, right: XQDA metric)]{
							\includegraphics[width=0.49 \linewidth]{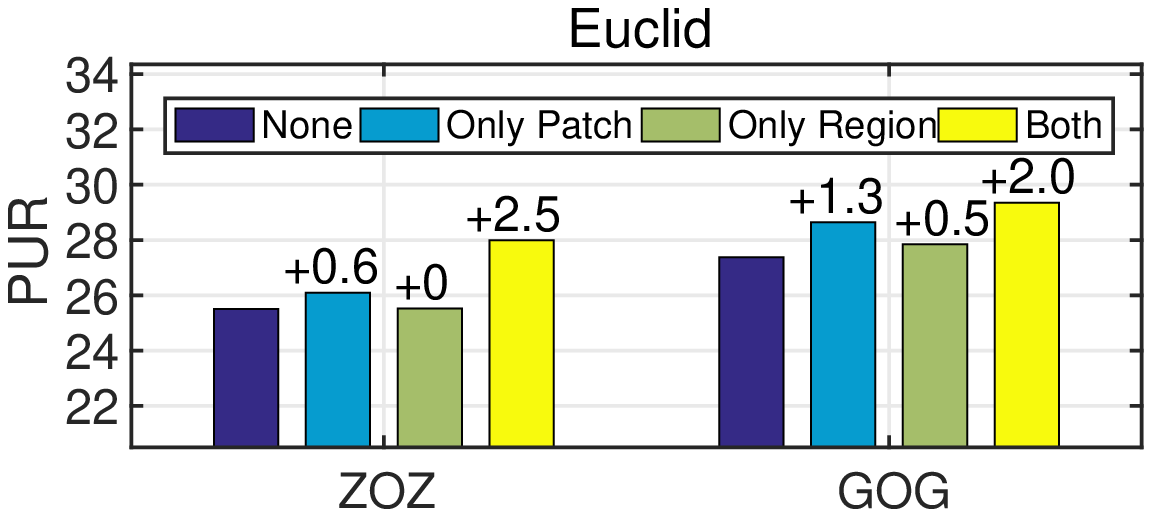}
							\includegraphics[width=0.49 \linewidth]{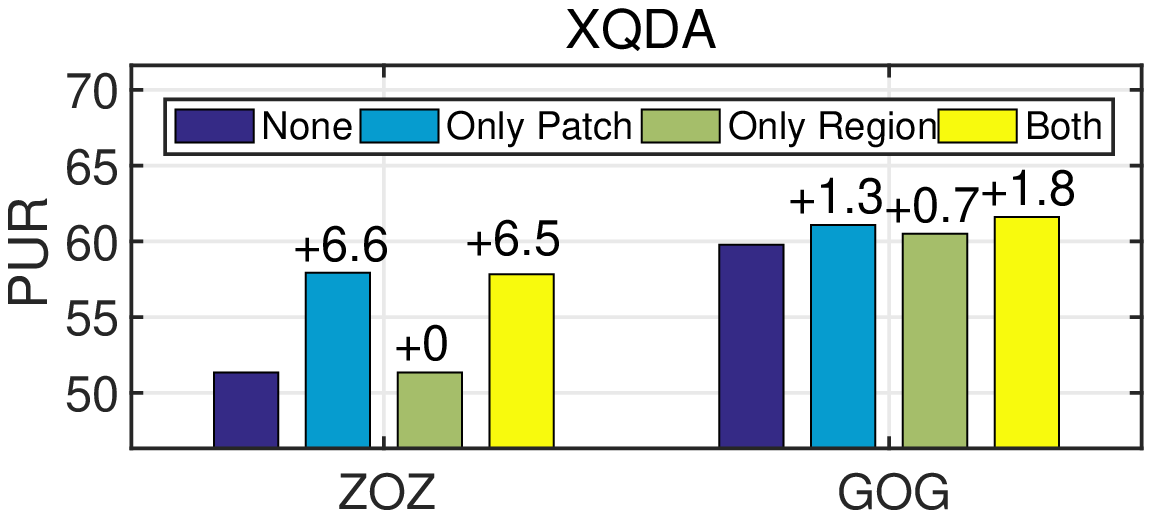}
							\label{fig:scale}}
					\end{minipage}
                                        \vspace{-2mm}
					\begin{minipage}[b]{1 \linewidth }
						\centering\subfigure[ Norm normalization (left: Euclidean distance, right: XQDA metric)]{
							\includegraphics[width=0.49 \linewidth]{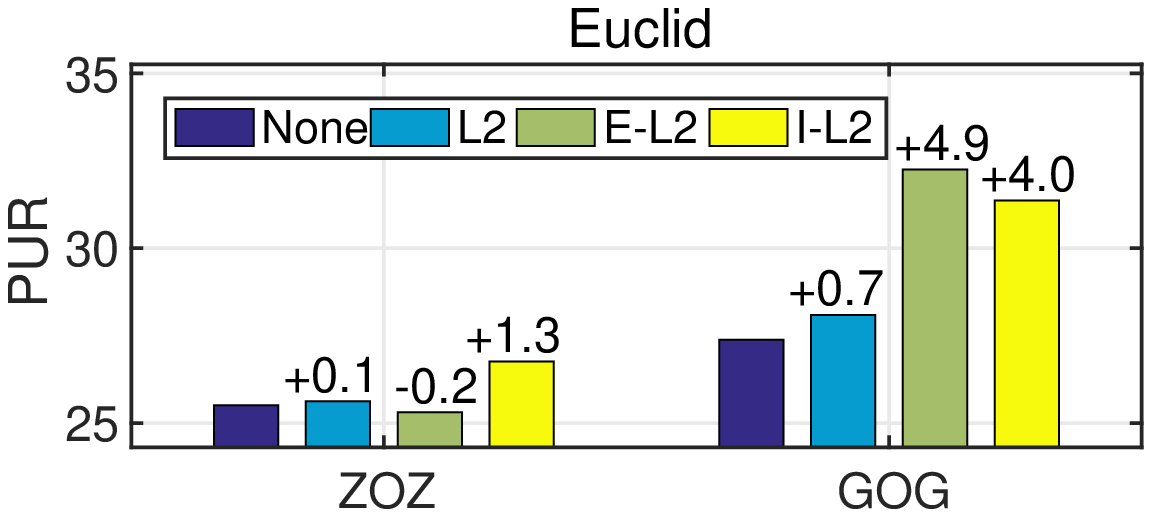}
							\includegraphics[width=0.49 \linewidth]{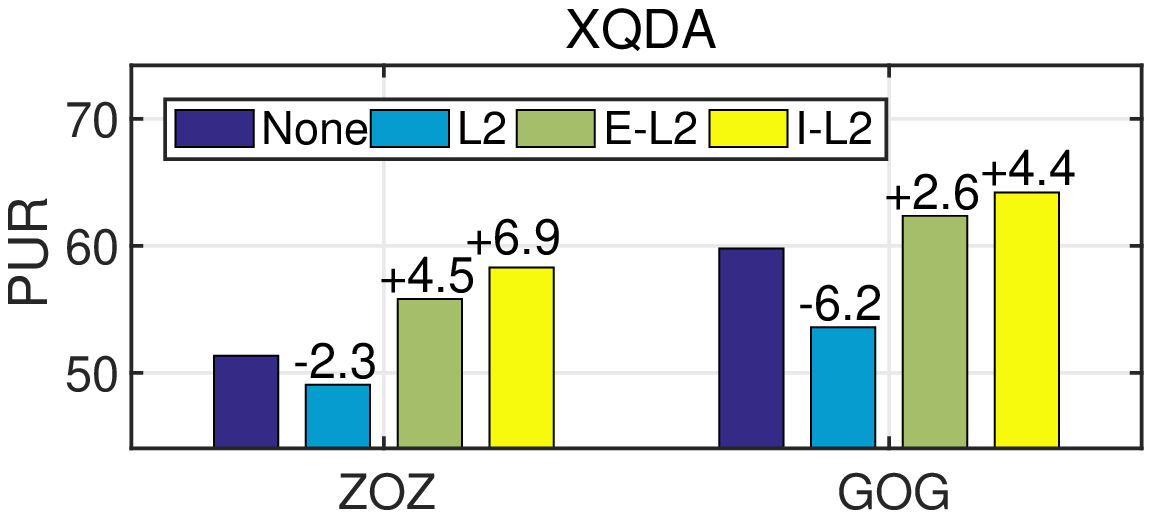}
							\label{fig:scale}}
					\end{minipage}
					\caption{ Evaluating the impact of the normalizations on the VIPeR dataset: (a) The scale normalization, (b) The norm normalization. The numbers in each figure show the performance gain of PUR by each normalization. }
					\label{fig:comparinorm}
					\vspace{-6mm}
				\end{center}
			\end{figure}

			\begin{table*}
				\caption{Evaluating the impact of the hierarchical Gaussian embedding with XQDA metric learning (CMC@rank-r/PUR/mAP). 
                                        The mark \checkmark indicates that the hierarchy is included.
                                        The best scores among \{each base embedding, all embeddings\} are indicated in \{\bc{blue}, \rc{red}\}, respectively.}
				\vspace{-6mm}
				\label{table:impact_hie}
				\begin{center}
					{ 
						\scriptsize
						\renewcommand{\arraystretch}{0.65} 
						\tabcolsep = 0.65 mm 
						\begin{tabular}{ |cc||ccccc||ccccc||ccccc||ccccc||ccccc| }
							\hline
						&                &    \multicolumn{5}{c||} {\bf VIPeR}  & \multicolumn{5}{c||}{\bf GRID}  & \multicolumn{5}{c||}{\bf CUHK01} & \multicolumn{5}{c||}{\bf CUHK03}  & \multicolumn{5}{c|}{\bf Market-1501} \\ 
						{\bf Methods}  &       {\bf Hierarchy}  &    {\bf r=1} & {\bf r=5} & {\bf r=10} & {\bf r=20} & {\bf PUR} &   {\bf r=1} & {\bf r=5} & {\bf r=10} & {\bf r=20} & {\bf PUR} &  {\bf r=1} & {\bf r=5} & {\bf r=10} & {\bf r=20} & {\bf PUR} &  {\bf r=1} & {\bf r=5} & {\bf r=10} & {\bf r=20} & {\bf PUR} &  {\bf r=1} & {\bf r=5} & {\bf r=10} & {\bf r=20} & {\bf mAP}\\ \hline
						Mean                  &                     & 20.7 & 45.9 & 59.1 & 72.0 & 36.4 & 8.5 & 21.3 & 29.6 & 37.5 & 24.6 &  22.6 & 45.9 & 56.8 & 67.5 & 37.1 & 26.9 & 57.9 & 71.5 & 83.5 & 33.2 & 30.6 & 52.0 & 61.6 & 70.9 & 13.6 \\ \hline 
						Cov                   &                     & 30.8 & 58.5 & 70.3 & 81.4 & 45.0 & 14.9 & 29.9 & 38.2 & 50.3 & 32.0 & 48.4 & 71.5 & 79.8 & 87.0 & 59.2 & 54.7 & 82.6 & 90.3 & 95.5 & 58.7 & 52.5 & 70.4 & 77.7 & 84.1 & 27.0 \\ 
						COC                   & \checkmark          & \bc{35.5} & \bc{67.5} & \bc{79.9} & \bc{88.8} & \bc{53.0} & \bc{18.1} & \bc{36.2} & \bc{47.0} & \bc{57.2} & \bc{37.4} & \bc{58.6} & \bc{80.1} & \bc{86.5} & \bc{92.3} & \bc{68.0} & \bc{80.0} & \rc{95.8} & \rc{98.0} & \rc{99.3} & \bc{80.5} & \bc{55.9} & \bc{77.0} & \bc{83.3} & \bc{88.6} & \bc{32.2} \\ \hline 
						ZmG                   &                     & 36.5 & 64.9 & 75.5 & 85.6 & 50.5 & 16.6 & 32.9 & 41.3 & 52.8 & 34.4 & 49.9 & 74.4 & 82.8 & 88.8 & 61.3 & 56.5 & 83.8 & 90.6 & 95.3 & 60.0 & 52.6 & 71.6 & 78.4 & 84.5 & 27.4 \\ 
						ZOZ                   & \checkmark          & \bc{48.9} & \bc{76.5} & \bc{86.2} & \bc{93.0} & \bc{61.6} & \bc{28.1} & \rc{49.9} & \rc{61.1} & \rc{71.5} & \rc{48.0} & \rc{71.3} & \bc{88.1} & \bc{92.8} & \bc{96.2} & \bc{77.0} & \rc{81.6} & \bc{95.5} & \bc{97.8}  & \bc{99.0} & \rc{81.0} & \bc{63.5} & \bc{81.5} & \bc{87.2} & \bc{91.3} & \bc{38.3} \\ \hline 
						Gauss                 &                     & 37.8 & 65.0 & 76.4 & 86.2 & 51.4 & 17.8 & 33.0 & 43.1 & 53.6 & 35.1 & 51.1 & 75.5 & 83.2 & 89.0 & 62.1 & 60.0 & 86.0 & 92.3 & 96.5 & 63.0 &  53.1 & 72.9 & 80.1 & 85.9 & 29.2  \\ 
						GOG                   & \checkmark          & \rc{50.9} & \rc{80.8} & \rc{89.4} & \rc{94.6} & \rc{64.6} & \rc{28.3} & \bc{49.3} & \bc{59.7} & \bc{71.0} & \bc{47.8} & \bc{70.5} & \rc{88.7} & \rc{94.0} & \rc{97.1} & \rc{77.5} & \bc{80.8} & \bc{95.5} & \bc{97.7}  & \bc{99.1} & \bc{80.8} & \rc{66.5} & \rc{82.7} & \rc{88.2} & \rc{92.2} & \rc{41.2} \\ \hline 
                                                \end{tabular}
					}
				\end{center}
				\vspace{-5mm} 
			\end{table*}
			\begin{table*}
				\caption{Evaluating the impact of the scale and norm normalizations with XQDA metric learning (CMC@rank-r/PUR/mAP). The mark \checkmark indicates that the scale normalization is used. The best scores among \{ZOZ,  GOG\} are indicated in \{\bc{blue}, \rc{red}\}, respectively.}
				\vspace{-6mm}
				\label{table:impact}
				\begin{center}
					{ 
						\scriptsize
						\renewcommand{\arraystretch}{0.65} 
						\tabcolsep = 0.6 mm 
						\begin{tabular}{ |ccc||ccccc||ccccc||ccccc||ccccc||ccccc| }
						\hline
           					&  &  &   \multicolumn{5}{c||} {\bf VIPeR}  & \multicolumn{5}{c||}{\bf GRID}  & \multicolumn{5}{c||}{\bf CUHK01} & \multicolumn{5}{c||}{\bf CUHK03}  & \multicolumn{5}{c|}{\bf Market-1501}  \\ 
						{\bf Methods } & {\bf Scale} & {\bf Norm}  &  {\bf r=1} & {\bf r=5} & {\bf r=10} & {\bf r=20} & {\bf PUR} &   {\bf r=1} & {\bf r=5} & {\bf r=10} & {\bf r=20} & {\bf PUR} &  {\bf r=1} & {\bf r=5} & {\bf r=10} & {\bf r=20} & {\bf PUR} &  {\bf r=1} & {\bf r=5} & {\bf r=10} & {\bf r=20} & {\bf PUR}  &  {\bf r=1} & {\bf r=5} & {\bf r=10} & {\bf r=20} & {\bf mAP}  \\ \hline
						\multirow{6}{*}{ZOZ}   &                   &               & 40.0 & 65.5 & 75.6 & 85.3 & 51.4 & 24.2 & 45.1 & 55.3 & 67.0 & 44.4 & 49.6 & 66.8 & 74.0 & 80.8 & 54.6 & 41.9 & 64.3 & 75.3 & 83.8 & 40.2 & 53.0 & 72.7 & 79.8 & 85.6 & 25.4 \\ 
                       							& \checkmark        &               & 45.0 & 73.3 & 82.9 & 90.6 & 57.6 & 25.7 & 46.2 & 58.6 & 69.7 & 46.1 & 47.7 & 66.2 & 73.8 & 81.3 & 54.3 & 20.0 & 42.5 & 57.5 & 72.7 & 22.1 & 35.4 & 53.8 & 61.8 & 68.8 & 12.2 \\ 
                       							&                   & E-L2          & 42.8 & 69.9 & 80.3 & 89.5 & 55.8 & 27.0 & 48.8 & 58.5 & 68.8 & 46.1 & 66.7 & 85.5 & 90.3 & 94.3 & 73.3 & 75.0 & 92.0 & 95.3 & 97.3 & 74.3 & 61.1 & 78.8 & 85.3 & 90.2 & 35.4 \\ 
                       							&                   & I-L2          & 46.2 & 72.6 & 82.1 & 91.1 & 58.3 & 27.0 & 47.6 & 59.3 & 70.1 & 46.4 & 66.2 & 85.6 & 91.5 & 95.3 & 73.6 & 80.9 & \bc{95.5} & 97.5 & \bc{99.0} & 80.3 & 60.6 & 78.7 & 85.0 & 89.9 & 34.9 \\ 
                      							& \checkmark        & E-L2          & 47.5 & 75.3 & 84.7 & 92.8 & 60.5 & 27.5 & 48.6 & 59.5 & 69.8 & 47.0 & 69.4 & 87.1 & 91.4 & 95.4 & 75.4 & 78.3 & 93.9 & 96.5 & 98.3 & 77.8 & 63.4 & 81.0 & 86.7 & \bc{91.6} & 37.8 \\ 
                      							& \checkmark        & I-L2          & \bc{48.9} & \bc{76.5} & \bc{86.2} & \bc{93.0} & \bc{61.6} & \bc{28.1} & \bc{49.9} & \bc{61.1} & \bc{71.5} & \bc{48.0} & \bc{71.3} & \bc{88.1} & \bc{92.8} & \bc{96.2} & \bc{77.0} & \bc{81.6} & \bc{95.5} & \bc{97.8} & \bc{90.0} & \bc{81.0} &  \bc{63.5} & \bc{81.5} & \bc{87.2} & 91.3 & \bc{38.3}  \\ \hline \hline
						\multirow{6}{*}{GOG}    &                   &               & 45.9 & 75.1 & 85.4 & 92.2 & 59.8 & 23.3 & 42.7 & 53.4 & 66.6 & 43.2 & 63.2 & 80.7 & 86.6 & 90.9 & 68.8 & 7.9 & 27.1 & 44.2 &  62.4 & 13.5 & 22.0 & 35.1 & 41.7 & 50.4 & 6.5 \\ 
                       							& \checkmark        &               & 47.9 & 77.1 & 86.9 & 93.4 & 61.6 & 24.0 & 44.1 & 55.8 & 68.1 & 44.4 & 64.1 & 82.5 & 87.7 & 92.1 & 70.3 & 10.6 & 34.8 & 51.0 & 68.3 & 17.3 & 20.8 & 35.8 & 43.4 & 52.9 & 6.7 \\ 
                       							&                   & E-L2          & 47.5 & 78.3 & 87.9 & 93.9 & 62.4 & 25.8 & 46.8 & 57.5 & 68.6 & 45.5 & 67.6 & 87.0 & 92.0 & 95.6 & 75.0 & 80.3 & 95.0 & 97.7 & 99.0 & 80.2 & 65.1 & 81.7 & 87.0 & 91.4 & 39.5 \\ 
                     							&                   & I-L2          & 50.5 & 80.3 & 89.1 & \rc{94.6} & 64.2 & \rc{28.6} & \rc{50.1} & \rc{60.8} & \rc{71.8} & \rc{48.0} & \rc{71.3} & \rc{89.0} & 93.6 & \rc{97.1} & \rc{77.9} & 80.2 & 95.3 & 97.8 & \rc{99.1} & 80.1 & 66.4 & 82.5 & 88.0 & 91.8 & 41.1 \\ 
                       							& \checkmark        & E-L2          & 49.7 & 79.7 & 88.7 & 94.5 & 63.6 & 24.8 & 47.0 & 58.4 & 68.9 & 46.0 & 67.3 & 86.9 & 91.8 & 95.9 & 74.9 & 80.5 & 95.5 & \rc{97.9} & 99.0 & 80.5 & 65.1 & 81.7 & 87.7 & 91.6 & 39.7 \\ 
                       							& \checkmark        & I-L2          & \rc{50.9} & \rc{80.8} & \rc{89.4} & \rc{94.6} & \rc{64.6} & 28.3 & 49.3 & 59.7 & 71.0 & 47.8 & 70.5 & 88.7 & \rc{94.0} & \rc{97.1} & 77.5 & \rc{80.8}& \rc{95.5} &  97.7 & \rc{99.1} & \rc{80.8} & \rc{66.5} & \rc{82.7} & \rc{88.2} & \rc{92.2} & \rc{41.2} \\ \hline
                                                \end{tabular}
					}
				\end{center}
				\vspace{-3mm} 
			\end{table*}

\noindent{\bf Impact of the scale normalization. }
We analyze the impact of the scale normalization in detail. Fig.~\ref{fig:comparinorm} (a) compares the PURs when the scale normalization is applied on either one of the patch or region Gaussians, and also on both of them. For this comparison, the norm normalization is not applied. 

The results show the followings:
The performance improvements by the scale normalization are larger on the patch Gaussians than on the region Gaussians for both ZOZ and GOG. This is because the discriminative ability in the intermediate representation is degraded, and is compensated for by the scale normalization.
These results support our analysis in $\S$\ref{sec:scale}: the hierarchical Gaussian matrices embed the outer product of the vectorized patch Gaussian matrices and the biased elements in the patch Gaussian matrices degrade the discriminative ability.
Although the improvement attributed to the normalization of the region Gaussians is insignificant, the highest scores are typically obtained by normalizing both the region and patch Gaussians.

\noindent{\bf Impact of the norm normalization. }
We compare the norm normalization in detail. Fig.~\ref{fig:comparinorm} (b) compares the PURs when the each of the standard L2, E-L2, and I-L2 normalizations is applied. For this comparison, the scale normalization is not applied.

From the results, we can see that:
(1) Both E-L2 and I-L2 normalizations improve the performance of the original features. 
Although the standard L2 normalization slightly improves the performance of the Euclidean metric, it decreases the performance of XQDA. 
These results confirm the effect of removing the large bias which exist especially in the descriptors.
(2) Intrinsic mean removal improves the performance more than extrinsic mean removal, which suggests that greater respect for the Riemannian geometry enables higher performance.

We summarize the impact of the scale and norm normalizations in Table~\ref{table:impact}. For the sake of generality, the table contains the results on all five datasets. Note that on the large datasets, \ie the CUHK03 and Market-1501 datasets, we observed unstable optimal dimension estimation of XQDA due to the large values of the GOG/ZOZ descriptors when the norm normalization is not applied.
Therefore, the scale normalization did not increase the performance. However, the norm normalization stabilizes the results and we can confirm the improvement of the scale normalization when the norm normalizations are applied.

\begin{figure}[t]
				\begin{center}
					\includegraphics[width=0.99 \linewidth]{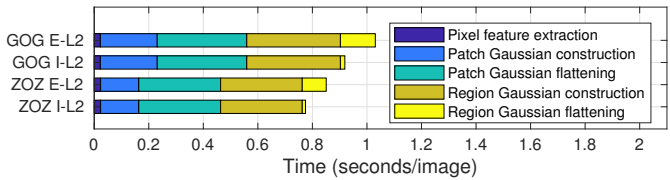}
					\vspace{-2mm}
					\caption{ Breakdown of computational time of feature extraction. }
					\label{fig:time}
					\vspace{-6mm}
				\end{center}
			\end{figure}

\def\subfigcapskip{-5pt}  
\noindent{\bf Running time. }
The HGDs are implemented in MATLAB\footnotemark with MEX functions for calculation of the covariance matrices, and run on a PC equipped with an Intel Xeon E5-2687W v3 @3.1 GHz CPU. The running times for extracting the descriptors are shown in Fig.~\ref{fig:time}. 
The listed times are the sum of the descriptors of the four-color spaces and the average of all images on the VIPeR dataset. Owing to its smaller dimensionality, ZOZ is about 1.2 times faster than GOG. Most of the time is spent to construct/flatten the patch Gaussians, and to construct the region Gaussians.
Since the number of the regions is small, the run time required to flatten the region Gaussians is small when E-L2 normalization is used.
The time for E-L2 normalization, \ie for mean removal and L2 normalization, is negligible (below 0.001 sec./image). 
For I-L2 normalization, the time for flattening the region Gaussian increases because it multiplies the matrix $\mathvc{M}^{-\frac{1}{2}}$ for every region Gaussian matrices. 
The matching time of ZOZ is also less than that of GOG owning to its smaller dimensionality.
\footnotetext{Will be released at \url{http://www.i.kyushu-u.ac.jp/~matsukawa/ReID.html}.}
			
			\begin{table*}[t]
\caption{Comparison of state-of-the-art descriptors with XQDA metric learning (CMC@rank-r/PUR/mAP).
The first, second, and third blocks respectively show the comparison among meta-descriptors, re-id descriptors, and combination of HGDs and re-id descriptors. 
In the first block, patch weights are not used and E-L2 normalization are applied. The best scores among each comparison are indicated in \bc{blue}, \rc{red}, and {\bf bold}, respectively.}
				\vspace{-6mm}
				\label{table:compari}
				\begin{center}
					{ 
						\scriptsize
						\renewcommand{\arraystretch}{0.65} 
						\tabcolsep = 0.5mm
						\begin{tabular}{ |cc||ccccc||ccccc||ccccc||ccccc||ccccc| }
							\hline
							&     & \multicolumn{5}{c||}{\bf VIPeR} & \multicolumn{5}{c||}{\bf GRID} & \multicolumn{5}{c||} {\bf CUHK01 }  & \multicolumn{5}{c||}{\bf CUHK03}  & \multicolumn{5}{c|}{\bf Market-1501} \\ 
							{\bf Methods} & {\bf Dim.} &  {\bf r=1} & {\bf r=5} & {\bf r=10} & {\bf r=20} &  {\bf PUR} &  {\bf r=1} & {\bf r=5} &  {\bf r=10} & {\bf r=20}  &  {\bf PUR} &  {\bf r=1} & {\bf r=5} &  {\bf r=10} & {\bf r=20} & {\bf PUR} &  {\bf r=1} & {\bf r=5} & {\bf r=10} & {\bf r=20} &  {\bf PUR} &  {\bf r=1} & {\bf r=5} & {\bf r=10} & {\bf r=20}  & {\bf mAP} \\ \hline 
     							GOLD~\cite{Serra15}       &  1,169 & 27.1 & 54.6 & 66.5 & 77.7 & 41.9 & 10.7 & 21.5 & 29.3 & 37.3 & 25.9 & 43.6 & 65.5 & 74.8 & 82.1 & 53.4 & 49.0 & 76.2 & 87.0 & 94.0 & 53.0 & 43.9 & 62.4 & 70.1 & 77.4 & 21.0 \\ 
     							HASC~\cite{Biagio}        &  1,904 & 30.9 & 58.6 & 70.6 & 81.8 & 46.0 & 12.6 & 27.0 & 35.5 & 47.3 & 31.0 & 47.5 & 70.2 & 78.4 & 85.3 & 57.4 & 58.0 & 83.8 & 91.3 & 96.2 & 61.2 & 46.9 & 66.3 & 74.9 & 82.7 & 24.3 \\ 
     							LDFV~\cite{MaSJ12}        &  6,944 & 26.8 & 55.7 & 69.4 & 82.1 & 44.5 & 15.6 & 34.1 & 44.7 & 55.3 & 36.4 & 46.0 & 71.5 & 80.1 & 87.0 & 58.1 & 64.7 & 88.9 & 93.1 & 96.5 & 66.5 & 42.7 & 62.9 & 72.3 & 80.8 & 21.5 \\ 
     							COC~\cite{serra14}        &  16,828 & 33.5  & 63.8 & 75.9  & 86.8 & 50.4 & 18.2 & 33.8 & 43.5 & 54.6 & 36.3 & 54.5 & 77.3 & 84.8 & 90.6 & 64.7 & 78.4 & 95.5 & 98.0 & 99.3 & 79.2 & 51.3 & 71.6 & 79.2 & 85.7 & 28.2 \\ 
     							R-VLAD~\cite{Faraki15}    &  30,464 &  28.1 & 58.1 & 71.7 & 84.0 & 46.6 & 12.8 & 30.9 & 41.5 & 52.2 & 33.9 & 58.4 & 80.6 &  88.0 & 93.0 & 68.3 & \bc{80.8} & \bc{96.5} & \bc{98.8} & \bc{99.4} & \bc{82.0} & 53.1 & 75.0 & 82.2 & 87.9 & 30.6 \\
     							FV-L$^2$EMG~\cite{Li16}   &  38,304 &  31.7 & 62.2 & 75.7 & 86.1 & 49.9 & 20.1 & 39.2 & 51.0 & 62.9 & 40.7 &  55.3 & 79.1 & 86.9  & 92.5 & 66.2 & 71.2 & 91.9 & 95.8 & 97.8 & 72.4 &  50.5 &  71.2 & 78.7 & 85.4  & 27.3  \\
     							{\bf ZOZ}                 &  16,828 &  44.7 & 75.2 & 85.6 & 92.6 & 59.6 & \bc{24.6} & 43.0 & \bc{53.9} & \bc{66.5} & 43.7 & \bc{65.9} & \bc{84.8} & 89.7 & 93.7& \bc{72.3}  & 75.0 & 93.3 & 96.1 & 98.0 & 75.0 &  56.5 & 76.6 & 83.9 & 89.5 & 33.1 \\
     							{\bf GOG}                 &  27,622 &  \bc{47.0} & \bc{78.6} & \bc{89.2} & \bc{94.8} & \bc{62.6} & \bc{24.6} & \bc{43.2} & 53.8 & 63.9 & \bc{43.9} & 65.0 & 84.7 & \bc{90.0} & \bc{94.3}& \bc{72.3} & 80.5 & 96.0 & 98.0 & 99.2 & 80.7 & \bc{59.4} & \bc{77.9} & \bc{84.6} & \bc{89.8} & \bc{35.5} \\ \hline \hline
     							gBiCov~\cite{ma2014} & 5,940 & 22.9 & 49.8 & 63.7 & 77.6 & 40.4 & 10.6 & 22.5 & 30.5 & 41.6 & 28.4 & 32.3 & 57.7 & 68.1 & 78.7 & 47.6 & 35.4 & 71.7 & 83.8 & 93.4 & 45.8 & 23.4 & 46.5 & 57.2 & 67.5 & 10.6 \\
     							CH+LBP~\cite{xiong} & 32,250 & 27.7 & 55.1 & 69.3  & 82.4 & 45.0 & 16.2 & 33.4 & 45.0 & 57.2 & 36.7 & 39.3 & 69.0  & 80.2 & 88.9 & 57.2 & 50.1 & 81.8 & 89.8 & 95.8 & 56.6 & 32.7 & 54.6 & 64.0 & 73.4 & 16.0 \\
    	 						LOMO~\cite{Liao15} & 26,960 & 41.2 & 71.2 & 82.4 & 91.3 & 56.9 & 17.8 & 38.5 & 46.3 & 56.6 & 36.7 & 62.4 & 85.0 & 90.7 & 94.8 & 72.0 & 64.9 & 91.4 & 95.7 & 98.3 & 69.2 & 40.0  &  64.3  & 73.4  & 81.4 & 21.0  \\
     							HIPHOP~\cite{Chen17} & 84,096 & 48.5 & 78.6 & 87.3 & 94.2 & 62.6 & 19.5 & 42.6 & 51.7& 62.0 & 41.3 &  64.5 & 86.2 & 92.2 & 96.1 & 73.6 & 59.3 & 86.5 & 92.6 & 97.3 & 63.4 & 48.1& 69.8 & 78.8 & 85.8 & 26.2\\	
                                                        FTCNN~\cite{Matsukawa16b} & 4,096  & 42.9 & 72.1 & 82.2 & 91.6 & 57.4 & 26.4 & 45.2 & 53.9 & 63.4 & 43.5 & 58.4 & 81.7 & 88.7 & 93.9 & 69.0 & 66.2 & 91.5 & 95.8 & 98.1 & 70.2 & 50.1 & 73.5 & 80.5 & 87.0 & 28.7  \\
     							{\bf ZOZ}  & 16,828 &  48.9 & 76.5 & 86.2 & 93.0 & 61.6 & 28.1 & 49.9 & 61.1 & 71.5 & 48.0 & \rc{71.3} & 88.1 & 92.8 & 96.2 & 77.0 & 81.6 & 95.5 & 97.8 & 99.0 & 81.0 & 63.5 & 81.5 & 87.2 & 91.3 & 38.3 \\ 
     							{\bf GOG}  & 27,622 &  50.9 & 80.8 & \rc{89.4} & 94.6 & 64.6 & 28.3 & 49.3 & 59.7 & 71.0 & 47.8 & 70.5 & \rc{88.7} & \rc{94.0} & \rc{97.1} & \rc{77.5} & 80.8 & 95.5 & 97.7 & 99.1 & 80.8 & 66.5 & 82.7 & 88.2 & 92.2 & 41.2 \\ 
     							{\bf HGDs(ZOZ+GOG)}   & 44,450 & \rc{52.2} & \rc{80.9} & 89.1 & \rc{94.8} & \rc{65.1} & \rc{28.6} & \rc{51.0} & \rc{61.8} & \rc{73.2} & \rc{48.7} & 71.1 &  88.6 & 93.1 & \rc{97.1} & \rc{77.5} & \rc{82.7} & \rc{96.0} & \rc{98.0} & \rc{99.3} & \rc{82.1}  & \rc{67.8} & \rc{84.5} & \rc{89.0} & \rc{93.0} & \rc{42.8} \\ \hline \hline
     							{\bf HGDs}+gBiCov &  50.390 & 52.3 & 80.9 & 89.1 & 94.9 & 65.1 &  28.6 & 51.0 & 61.6 & 73.3 & 48.7 & 71.4 & 88.7 & 93.3 & 97.2 & 77.7 & 82.8 & 96.0 & 98.0 & {\bf 99.3} & 82.1 & 67.9 & 84.1 & 89.1 & 93.4 & 42.9 \\
                                                        {\bf HGDs}+CH+LBP &  76,700 & 52.2 & 81.3 & 89.6 & 95.2 & 65.3 &  28.6 & 51.5 & 63.0 & 73.8 & 49.0 & 72.2 & 88.3 & 93.7 & 97.4 & 78.1 & 83.0 & 95.6 & 98.0 & {\bf 99.3} & 82.3 & 67.0 & 83.9 & 89.0 & 92.5 & 42.7 \\
                                                        {\bf HGDs}+LOMO   &  71,410 & 53.5 & 82.0 & 90.2 & 95.3 & 66.3 &  29.3 & 52.2 & 62.7 & 73.0 & 49.1 & 74.4 & 89.9 & 94.3 & 97.8 & 79.9 & 83.5 & 95.8 & 98.2 & {\bf 99.3} & 82.8  & 68.3 & 84.9 & 89.6 & 93.4 & 44.1 \\ 
                                                        {\bf HGDs}+HIPHOP & 128,546 & 53.4 & 81.6 & 89.8 & 95.3 & 66.0 &  28.6 & 52.1 & 62.2 & 73.2 & 48.9 & 73.9 & 89.9 & 94.2 & 97.6 & 79.5 & 82.8 & 96.2 & 98.0 & {\bf 99.3} & 82.3  & 68.1 & 83.7 & 89.5 & 93.2 & 43.8 \\
     							{\bf HGDs}+FTCNN  &  48,546 & {\bf 54.7} & {\bf 83.4} & {\bf 90.9} & {\bf 95.7} & {\bf 67.2} & {\bf 30.6} & {\bf 52.6} & {\bf 63.9} & {\bf 74.0} & {\bf 50.3} & {\bf 75.3} & {\bf 91.0} & {\bf 94.9} & {\bf 98.0} & {\bf 80.7} & {\bf 84.7} & {\bf 96.8} & {\bf 98.5} & {\bf 99.3} & {\bf 84.1} & {\bf 70.2} & {\bf 85.5} & {\bf 90.0} & {\bf 93.9} & {\bf 46.0} \\ \hline
                                                        \end{tabular}
					}
				\end{center}
				\vspace{-3mm} 
			\end{table*}
\vspace{-2mm} 
\subsection{Performance Comparison}
We compare the following aspects of the proposed HGDs with other state-of-the-art approaches: (1) Meta-descriptors; (2) Person re-id descriptors; (3) Complement of existing re-id descriptors; (4) State-of-the-arts on person re-id.
		
\noindent{\bf Comparison with existing meta-descriptors.}
We compare the HGDs with existing meta-descriptors: Heterogeneous Auto-Similarities of Characteristics (HASC)~\cite{Biagio}, Local Descriptors encoded by Fisher Vector (LDFV)~\cite{MaSJ12}, GOLD~\cite{Serra15}, R-VLAD~\cite{Faraki15}, and FV-L$^2$EMG~\cite{Li16}. 
HASC is composed of the covariance descriptor and the Entropy and Mutual Information (EMI) descriptor which captures the nonlinear dependency within pixel features. 
GOLD describes an image region by the mean vector and covariance matrix. The principal matrix logarithm and half-vectorization is applied to the covariance matrix to obtain the vector of covariance. Subsequently, the vectors the mean and covariance are concatenated into a feature vector.
LDFV encodes pixel features using FV coding, which encodes the difference of pixel features from pre-trained GMM means. By following the recommended setting~\cite{MaSJ12}, we set the number of GMM components to 16.
R-VLAD is the VLAD coding on the Riemannian manifold data. We used the Stein divergence~\cite{Sra12} for metric to encode local covariance matrices. We set the number of codebook to 32. FV-L$^2$EMG is a descriptor for which the patch Gaussian matrices are summarized by FV. We set the GMM components to 16. We did not observe a large difference in the performance of FV-L$^2$EMG when different GMM components were used.

We focus on the summarizing process of pixel features only, and discard other options, such as the spatial pyramid in GOLD. We extract each of the meta-descriptors from the same regions as HGDs and concatenate them. As well as HGDs, we use the fusion approach that concatenates the meta-descriptors extracted from the same pixel features of the four color spaces. For a fair comparison, we commonly use XQDA metric learning and apply the mean removal and L2 norm normalization (equivalent to E-L2 for SPD matrix descriptors). We do not use the patch weight for all descriptors since several descriptors do not support this weight.

We list the performances of the compared methods in the first row block of Table~\ref{table:compari}. The hierarchical descriptors, COC, R-VLAD and FV-L$^2$EMG clearly outperform the descriptors based on single-layered distribution (GOLD, HASC, and LDFV). The ZOZ and GOG outperform COC and R-VLAD, since ZOZ and GOG include the mean information, which is absent from the local covariance matrices used in COC and R-VLAD. We also see that ZOZ and GOG outperform FV-L$^2$EMG. This may be because the FV coding assumes diagonal covariance and thus the correlations within the local Gaussian vectors are not encoded.

\noindent{\bf Comparison with existing re-id descriptors.}
We compare HGDs with several state-of-the-art person descriptors used in supervised person re-id: gBiCov~\cite{ma2014}, Color Histogram(CH)+LBP~\cite{xiong}, LOMO~\cite{Liao15}, Fine-Tuned(FT) CNN~\cite{Matsukawa16b}, and Histogram of Intensity Pattern and Ordinal Pattern (HIPHOP)~\cite{Chen17}.
For these descriptors, we use the source codes provided by the authors. The default parameters of the codes are used for LOMO and gBiCov. Xiong \etal~\cite{xiong} conducted experiments using different region numbers to extract 28 bin CH and 2 uniform LBPs. Among them, we used 75 regions, which was the best setting. FTCNN conducts a fine-tuning of the pre-trained AlexNet~\cite{krizhevsky} using a pedestrian attribute dataset and extracts features from the FC6 layer. HIPHOP is a concatenation features of two histogram patterns extracted from lower layers of the pre-trained AlexNet. The purpose of this comparison is to show the superiority of the HGDs against the existing re-id descriptors. Hence, we use the final model for HGDs. We apply the XQDA metric learning for all descriptors.

The results are presented in the second block of Table~\ref{table:compari}.
Although LOMO and CH+LBP use a larger number of spatial regions and higher dimensional pixel features, both ZOZ and GOG outperform these descriptors by a large margin. The superiority of these descriptors originate from their hierarchal use of the mean and covariance information of pixel features, whereas LOMO uses only the mean information.
We also confirm that the ZOZ and GOG outperform CNN features such as HIPHOP and FTCNN. The results also confirm the fusion of ZOZ and GOG slightly increases the performance of GOG in most of the datasets.

\noindent{\bf Complement of existing re-id descriptors.}
In person re-id, it is customary to combine different features due to the difficulties of distinguishing similar persons. The proposed HGDs are expected to be complementary to other feature descriptors because HGDs extract only $8$-dimensional features from each pixel (for each descriptor in four color spaces), \eg the local edge information is only $4$ dimensional. Other descriptors extract richer edge patterns, \eg the LBP patterns in LOMO and convolutional filters in CNN features.
To investigate our expectation, we concatenate HGDs with the existing re-id descriptors. 

The results are presented in the third block of Table~\ref{table:compari}. The combinations of HGDs with LOMO or CNN features produce higher re-id scores than HGDs alone. FTCNN extracts features from more upper layer of trained-CNN than HIPHOP, yielding higher-level features. Interestingly, HGDs combined with FTCNN improves the re-id scores more than when combined with HIPHOP.

\noindent{\bf Comparison with state-of-the-arts.}
We compare the performance of HGDs with state-of-the-art person re-id approaches in Table~\ref{table:sota}. The proposed HGDs with XQDA metric learning is termed HGDs-XQDA. Since the combination of HGDs with FTCNN performed the best, we also use this method, which is termed HGDs-XQDA(+FTCNN).

On the VIPeR and GRID datasets, HGDs-XQDA significantly outperforms Gated-SCNN~\cite{Varior16a}, TCP~\cite{Cheng16}, DNS(Fusion)~\cite{Zhang16}, Fused-CNN~\cite{Subramaniam16}, and MESP~\cite{Chen2017b} in rank-1 rates (52.2\% and 28.6\% on the respective datasets). These results confirm that HGDs achieve accurate re-id results even on small sampled datasets where a sufficient amount of training samples is not available to train deep models. HGDs-XQDA(+FTCNN) surpass the most of sate-of-the-art approaches including SCSP~\cite{Chen16} and CRAFT-MFA(+LOMO)~\cite{Chen17} in rank-1 rates (54.7\% and 30.6\% on the respective datasets).

On the CUHK01 and CUHK03 datasets, HGDs-XQDA outperforms TCP, and DNS(fusion) and performs comparable to Gated-SCNN in rank-1 rates (60.8\%/71.1\% and 68.1\%/82.7\% in the single-/multi-shot settings on the respective datasets). 
HGDs-XQDA(+FTCNN) achieves 65.9\%/75.3\% and 71.4\%/84.7\% rank-1 rates on the respective datasets in the single-/multi-shot settings. 
These scores are comparable to those of Fused-CNN and are slightly below than those of CRAFT-MFA(+LOMO). Note that CRAFT-MFA(+LOMO) used HIPHOP+LOMO features that are of 2.3 times higher dimensional than HGDs+FTCNN. 
In addition, the choice of feature descriptors is independent from the CRAFT approach, therefore HGDs could be combined with this approach.

On the Market-1501 dataset\footnotemark, HGDs-XQDA achieves 67.8\%/76.2\% rank-1 rates and 42.8\%/52.9\% mAP in the single-/multi-query settings, respectively. 
These scores significantly higher than those of SCSP, MESP, DNS(Fusion), and slightly higher than those of Gated-SCNN. HGDs-XQDA(+FTCNN) achieves 70.2\%/78.6\% rank-1 rates and 46.0\%/56.1\% mAP for the single-/multi-query settings, respectively.  These scores are clearly higher than those of the Gated-SCNN and are comparable to those of CRAFT-MFA(+LOMO).
~\footnotetext{ 
Recently, the ResNet deep learning architecture is improving the performance on the Market-1501 dataset~\cite{Zheng16} (Unpublished papers are excluded in Table~\ref{table:sota}). We confirmed that the combination of HGDs with ResNet-50-IDE baseline~\cite{Zheng16} produces 81.7\% rank-1 rate and 61.7\% mAP in the single query-setting while those of ResNet alone were 77.4\% rank-1 rate and 56.2\% mAP.
}

			\begin{table}[t]
				\caption{State-of-the-art results (CMC@rank-r/mAP). The best scores are shown in {\bf bold}. The mark $^*$ indicates deep learning methods.}
				\vspace{-6mm}
				\label{table:sota}
			
				\begin{center}
					{ 
				                \scriptsize
						
						\renewcommand{\arraystretch}{0.65} 
						
						{\small (a) VIPeR} \\  \tabcolsep = 0.7mm
						\begin{tabular}{ |cc||cccc| }
							\hline\
							{\bf Methods} & {\bf Reference} & {\bf r=1} & {\bf r=5} & {\bf r=10 } & {\bf r=20}  \\ \hline 
							Gated-SCNN$^*$ & ECCV2016~\cite{Varior16a} & 37.8 & 66.9 & 77.4 & - \\
							TCP$^*$   & CVPR2016~\cite{Cheng16} & 47.8 & 74.7 & 84.8 & 89.2 \\ 
							DNS(Fusion)     & CVPR2016~\cite{Zhang16}   & 51.2 & 82.1 & 90.5 & 95.9 \\
							SCSP            & CVPR2016~\cite{Chen16}& 53.5 & 82.6 & {\bf 91.5} & 96.7 \\
							CRAFT-MFA(+LOMO)& PAMI2017~\cite{Chen17} & 54.2 & 82.4 & {\bf 91.5} & {\bf 96.9} \\ 
							{\bf HGDs-XQDA} & {\bf Ours} & 52.2 & 80.9 & 89.1 & 94.8  \\
							{\bf HGDs-XQDA(+FTCNN)}  & {\bf Ours} & {\bf 54.7} & {\bf 83.4} & 90.9 & 95.7 \\ \hline    
						\end{tabular}
                                                \vspace{1mm}
					        \\{\small (b) GRID} \\
						\begin{tabular}{ |cc||cccc| }
							\hline\
							{\bf Methods} & {\bf Reference} & {\bf r=1} & {\bf r=5} & {\bf r=10 } & {\bf r=20}  \\ \hline 
							Fused-CNN$^*$ & NIPS2016~\cite{Subramaniam16} & 19.2 & 38.4 & 53.6 & 66.4 \\
							MESP         & IJCV2017~\cite{Chen2017b} & 23.5 & 42.3 & 52.4 & 62.2 \\
							SCSP         & CVPR2016~\cite{Chen16} & 24.2 & 44.6 & 54.1 & 65.2 \\
							CRAFT-MFA(+LOMO)& PAMI2017~\cite{Chen17} & 26.0 & 50.6  & 62.5  & 73.3 \\
							{\bf HGDs-XQDA} & {\bf Ours} & 28.6 & 51.0 & 61.8 & 73.2 \\
							{\bf HGDs-XQDA(+FTCNN)}    & {\bf Ours} & {\bf 30.6} & {\bf 52.6} & {\bf 63.9} & {\bf 74.0} \\ \hline
						\end{tabular}
                                                \vspace{1mm}
						\\{\small (c) CUHK01} \\\tabcolsep = 0.3mm
						\begin{tabular}{ |cc||cccc||cccc| }
							\hline\
							&     &  \multicolumn{4}{c||}{\bf Single-shot} & \multicolumn{4}{c|}{\bf Multi-shot}  \\ 
							{\bf Methods} & {\bf Reference} & {\bf r=1} & {\bf r=5} & {\bf r=10 } &  {\bf r=20} & {\bf r=1} & {\bf r=5}  & {\bf r=10}  & {\bf r=20} \\ \hline 
							
							TCP$^*$   & CVPR2016~\cite{Cheng16} & 53.7 & 84.3 & 91.0 & 96,3 & - & - & - & - \\
							Fused-CNN$^*$   & NIPS2016~\cite{Subramaniam16} & 65.0 & - & 89.8 & 94.5 & - & -& - & - \\
							CRAFT-MFA(+LOMO)& PAMI2017~\cite{Chen17} & - & - & - & - & {\bf 78.8} & {\bf 92.6} & {\bf 95.3} & 97.8 \\
							{\bf HGDs-XQDA} & {\bf Ours} & 60.8 & 81.7 & 87.5 & 93.1 & 71.1 & 88.6 & 93.1& 97.1 \\
							{\bf HGDs-XQDA(+FTCNN)}  & {\bf Ours} & {\bf 65.9} & {\bf 84.6} & {\bf 90.3} & {\bf 95.1} & 75.3 & 91.0 & 94.9 & {\bf 98.0} \\ \hline
						\end{tabular}
                                                \vspace{1mm}
						\\{\small (d) CUHK03} \\ \tabcolsep = 0.3mm
						\begin{tabular}{ |cc||cccc||cccc| }
							\hline\
                                                        	&     &  \multicolumn{4}{c||}{\bf Single-shot} & \multicolumn{4}{c|}{\bf Multi-shot}  \\ 
							{\bf Methods} & {\bf Reference} & {\bf r=1} & {\bf r=5} & {\bf r=10 } & {\bf r=20} & {\bf r=1} & {\bf r=5}  & {\bf r=10} & {\bf r=20} \\ \hline 
							DNS(Fusion)       & CVPR2016~\cite{Zhang16}  & 54.7 & 84.8 & 94.8 & -  & - & - & - & -  \\
							Gated-SCNN$^*$        & ECCV2016~\cite{Varior16a} &  68.1 &  88.1 &  94.6 & -& - & - & - & - 　 \\ 
							Fused-CNN$^*$          & NIPS2016~\cite{Subramaniam16} & {\bf 72.0} & - & {\bf 96.0} & {\bf 98.3} & - & - & - & -  \\ 
                                                        CRAFT-MFA(+LOMO)       & PAMI2017~\cite{Chen17} & - & -& - & - & {\bf 87.5} & {\bf 97.4} & {\bf 98.7} & -   \\ 
                                                       	{\bf HGDs-XQDA}        & {\bf Ours}   & 68.1 & 90.8 & 95.2 & 97.7 & 82.7 & 96.0 & 98.0 & {\bf 99.3}  \\
							{\bf HGDs-XQDA(+FTCNN)}  & {\bf Ours} &  71.4 & 92.3 & 95.7 & 97.9 & 84.7 & 96.8 &  98.5 & {\bf 99.3} \\ \hline 
						\end{tabular}
                                                \vspace{1mm}
						\\{\small (e) Market-1501} \\ \tabcolsep = 1.5mm
						\begin{tabular}{ |cc||cc||cc| }
							\hline\
							&     &  \multicolumn{2}{c||}{\bf Single-query} & \multicolumn{2}{c|}{\bf Multi-query}  \\ 
							{\bf Methods} & {\bf Reference} & {\bf r=1} & {\bf mAP } & {\bf r=1} & {\bf mAP} \\ \hline 
							SCSP      & CVPR2016~\cite{Chen16} & 51.9 & 26.4 &  - & -  \\
							MESP      & IJCV2017~\cite{Chen2017b} & 53.1 & 26.7 &  - & -  \\
							DNS(Fusion) & CVPR2016~\cite{Zhang16} & 61.0 & 35.7  & 71.6 & 46.0  \\
							Gated-SCNN$^*$ & ECCV2016~\cite{Varior16a}  & 65.9 & 39.6  &  76.0 & 48.5\\ 
							CRAFT-MFA(+LOMO)   & PAMI2017~\cite{Chen17} & {\bf 71.8}  & 45.5 &{\bf 79.7} & 54.3 \\ 
							{\bf HGDs-XQDA} & {\bf Ours} &  67.8 & 42.8 &  76.2 & 52.9 \\
							{\bf HGDs-XQDA(+FTCNN)} & {\bf Ours} & 70.2 & {\bf 46.0} & 78.6 & {\bf 56.1} \\ \hline
				   	     \end{tabular}
					}
				\end{center}
				\vspace{-3mm} 
			\end{table}
%
\section{Conclusions}
We have proposed novel hierarchical meta-descriptors for person re-id. 
The proposed descriptors model both the mean and covariance information of pixel features in each of the patch and region hierarchies. 
Extensive experiments were conducted to confirm the importance of both the mean information of pixel features and the hierarchal distribution for person re-id.

We also verified that the scale normalization of Gaussian matrices enhances the performance of the hierarchical descriptors. Based on the normalization, we proposed zero mean Gaussian embedding, which achieves similar performance to the original Gaussian embedding with smaller dimensionality. 
Additionally, we investigated the feature norm normalization of SPD matrix-based descriptors. We showed that normalization with the intrinsic statistics of the Riemannian manifold enables us to obtain more accurate re-id results.

The normalization results have encouraged us to develop metric learnings of the SPD matrix descriptor for person re-id, \eg~\cite{HuangWSLC15, Harandi17}. In this regard, we plan to release the HGDs in matrix form. 
Another exciting future direction is to combine the hierarchical Gaussian embedding with CNN, \eg by using convolutional filters for input and by optimizing it with backpropagation~\cite{Yu17}.
Another possible direction would be to extend HGDs to video-based person descriptors.

%
%
%
		\ifCLASSOPTIONcaptionsoff
		\newpage
		\fi
%
%
%
%
%
		\bibliographystyle{IEEEtran}
		\bibliography{egbib}

\end{document}